\documentclass[preprint]{elsarticle}
\usepackage{geometry}
\usepackage{caption}
\usepackage{longtable}
\usepackage{rotating}
\usepackage{makecell}
\usepackage{rotating}
\usepackage{framed,multirow}
\usepackage{multirow,booktabs}
\usepackage[edges]{forest}
\usepackage{tikz}
\usetikzlibrary{trees}
\usepackage{placeins}
\usepackage{tablefootnote} 
\usepackage{amssymb}
\usepackage{amsmath}
\usepackage{mathtools}
\usepackage{latexsym}
\usepackage{multicol}
\usepackage{longtable}
\usepackage{booktabs}   
\usepackage{footnote}
\usepackage{svg}
\newcolumntype{Y}{>{\arraybackslash}X}
\usepackage[tables]{ragged2e}
\usepackage{capt-of}
\usepackage{subfigure}
\newcommand{\RA}[1]{{\color{black}{#1}}} 
\newcommand{\RB}[1]{{\color{black}{#1}}}   
\usepackage{lipsum}
\usepackage{url}
\usepackage{xcolor}
\usepackage{float}
\usepackage{array}
\usepackage{hyperref}
\usepackage{adjustbox}
\newcolumntype{P}[1]{>{\raggedright\arraybackslash}p{#1}}
\usepackage{threeparttable}
\makesavenoteenv{tabular}
\makesavenoteenv{table*}
\usepackage{xcolor}
\tikzset{parent/.style={align=center,text width=3cm,rounded corners=3pt},
subparent/.style={align=center,text width=3cm,rounded corners=3pt},
    child/.style={align=center,text width=8cm,rounded corners=3pt}
    }

\usepackage{tikz}
\def\checkmark{\tikz\fill[scale=0.2](0,.35) -- (.25,0) -- (1,.7) -- (.25,.15) -- cycle;}
\makeatletter
\def\ps@pprintTitle{%
  \let\@oddhead\@empty
  \let\@evenhead\@empty
  \def\@oddfoot{\reset@font\hfil\thepage\hfil}
  \let\@evenfoot\@oddfoot
}
\makeatother
\begin{document}
\urlstyle{same}

\begin{frontmatter}

\title{Segmentation in large-scale  cellular electron microscopy with deep learning: A literature survey}%

\author[1,2]{Anusha Aswath\corref{cor1}}
\cortext[cor1]{Corresponding author: 
  Tel.: +0-000-000-0000;  
  fax: +0-000-000-0000;}
  \ead{a.aswath@rug.nl}
\author[2]{Ahmad Alsahaf}
\author[2]{Ben N. G. Giepmans}

\author[1]{George Azzopardi}

\address[1]{Bernoulli Institute of Mathematics, Computer Science and Artificial Intelligence, University Groningen, Groningen, The Netherlands}
\address[2]{Dept. Biomedical Sciences of Cells and Systems, University Groningen, University Medical Center Groningen, Groningen, The Netherlands}

\begin{abstract}

Automated and semi-automated techniques in biomedical electron microscopy (EM) enable the acquisition of large datasets at a high rate. Segmentation methods are therefore essential to analyze and interpret these large volumes of data, which can no longer completely be labeled manually. In recent years, deep learning algorithms achieved impressive results in both pixel-level labeling (semantic segmentation) and the labeling of separate instances of the same class (instance segmentation). In this review, we examine how these algorithms were adapted to the task of segmenting cellular and sub-cellular structures in EM images. The special challenges posed by such images and the network architectures that overcame some of them are described. Moreover, a thorough overview is also provided on the notable datasets that contributed to the proliferation of deep learning in EM. Finally, an outlook of current trends and future prospects of EM segmentation is given, especially in the area of label-free learning.
\end{abstract}

\end{frontmatter}

\section{Introduction}

Electron microscopy (EM) is widely used in life sciences to study tissues, cells, subcellular components and (macro) molecular complexes at nanometer scale.~Two-dimensional (2D) EM aids in diagnosis of diseases, but routinely it still depends upon biased snapshots of areas of interest.~Automated pipelines for collection, stitching and open access publishing of 2D EM have been pioneered for transmission EM (TEM) images \citep{faas2012virtual} \RB{as well as scanning TEM (STEM)} \citep{sokol2015large} for acquisition of areas up to 1mm$^2$ at nanometer-range resolution. Nowadays, imaging of large areas at high resolution is entering the field as a routine method and is provided by most EM manufacturers. \RB{We term this nanotomy, for nano-anatomy}\citep{ravelli2013destruction,de2020large,dittmayer2021preparation}. The large-scale images allow for open access world-wide data sharing; see nanotomy.org\footnote{\url{www.nanotomy.org}} for  more than 50 published studies and the accessible nanotomy data. 

\begin{figure*}[t]
\centering
\includegraphics[trim={1.1cm 1.1cm 1.6cm 1.3cm},clip,width=\textwidth,keepaspectratio]{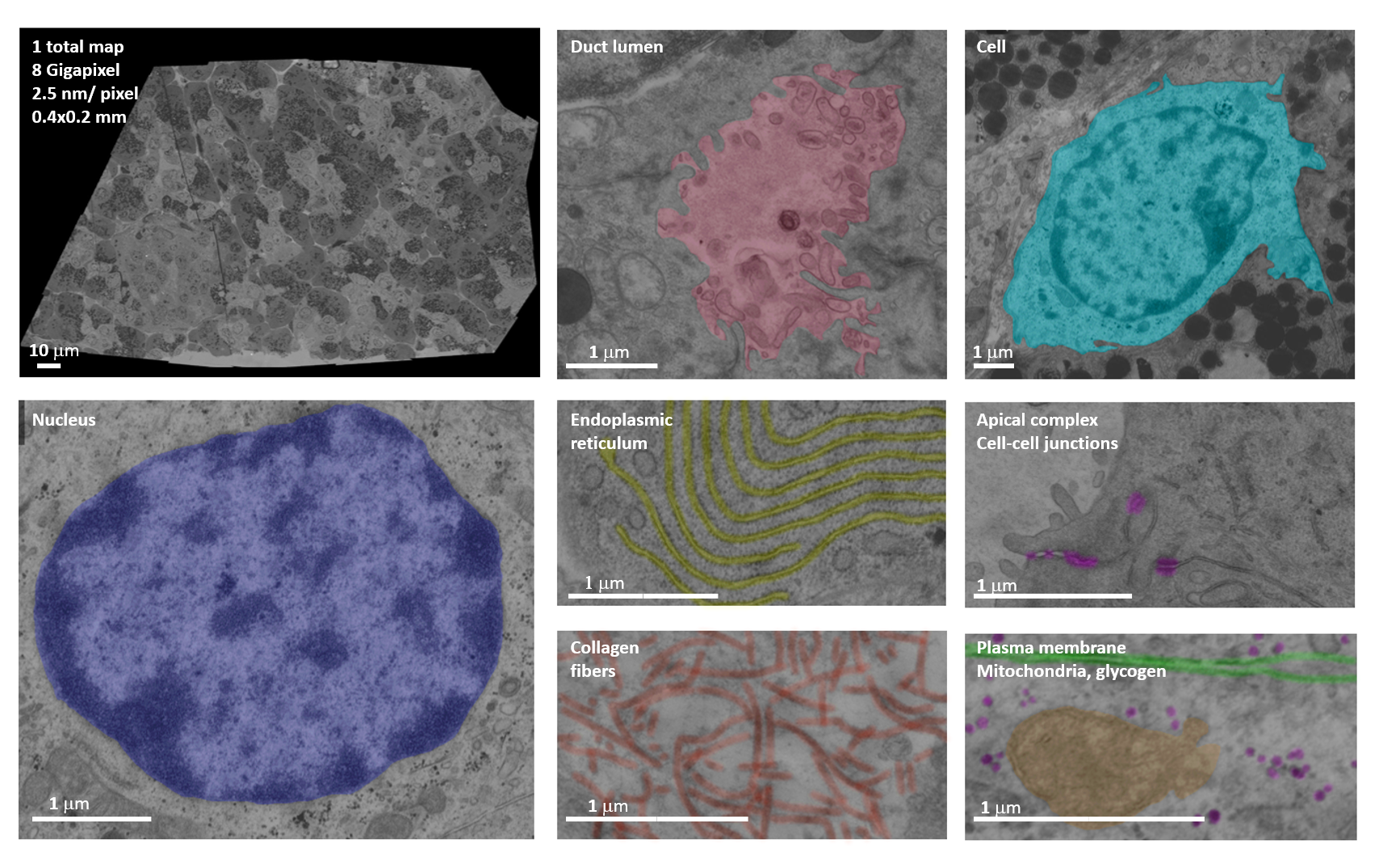}
    \caption{\RA{Large-scale EM (‘nanotomy’) of a section of human pancreas. Overview of a single large-scale EM (top-left) and snapshots from this total map at higher zoom showing several cellular, subcellular and macromolecular structures as indicated and annotated. Note the information density of these maps: millions of subcellular structures of a kind can be present per dataset \citep{de2020large}. Full access to digital zoomable data at full resolution is via \url{http://www.nanotomy.org}.}}
    \label{fig:Fig1}
\end{figure*}
A typical nanotomy dataset has a size of 5-25GB at 2.5nm pixel size. Nanotomy allows scientists to pan and zoom through different tissues or cellular structures, Fig.~\ref{fig:Fig1}.~Large-scale 2D EM provides unbiased data recording to discover events such as pathogenesis of diseases and \RB{morphological (shape and texture) changes at the subcellular level}. Moreover, nanotomy allows for the quantification of subcellular hallmarks.~With state-of-the-art 2D EM technology, such as multibeam scanning EM \citep{eberle2015high, ren2016transmission}, up to 100 times faster acquisition and higher throughput allows for imaging of tissue-wide sections in the range of hours instead of days. For a side-by-side example of single beam versus multibeam nanotomy, see  \citet{de2021state}. Given the automated and faster image acquisition in 2D EM a data avalanche (petabyte range per microscope/month) is becoming a reality.

Automated large-scale three-dimensional (3D) or volume EM (vEM), which creates stacks of images, is also booming \citep{peddie2014exploring,titze2016volume,peddie2022volume}.~The faster acquisition of 3D EM for serial-sectioning transmission EM (ssTEM) and serial block-face scanning EM (SBF-SEM) technologies can also lead to accumulation of petabytes of data. \RB{For instance, a complete brain volume of an adult fruit fly was imaged by ssTEM \citep{zheng2018complete}, which covered a single neuron cell in a volume of 1mm$^{3}$ or $10,000$ voxels and required 100 TB. Additionally, manual annotation is not practical due to the size of 3D EM datasets. An example by \citet{heinrich2021whole} shows that one person needed two weeks to manually label a fraction (1 $\mu m^3$) of a whole-cell volume containing tons of instances of various types of organelles, whereas the whole cell could take 60 person-years.} \RA{Whole-cell cryo-electron tomography (cryo-ET) has also advanced the capabilities of 3D EM to investigate the structure of cellular architecture and macromolecular assemblies in their native environment.}~The core data acquisition techniques of such 2D and 3D EM technologies are listed in Table \ref{tab:Table1}.

This increase in the scale and acquisition speeds of EM data accelerated the development of compatible methods for automatic analysis, especially in the areas of semantic and instance segmentation. Semantic segmentation classifies the pixels of an image into semantically meaningful categories, e.g. nuclei and background, while instance segmentation focuses on separating individual instances within the same class; e.g. the delineation of apposed mitochondria.

In the past, traditional image analysis methods as well as shallow learning algorithms\footnote{Shallow learning in this context refers to supervised machine learning with hand-crafted features, or traditional unsupervised techniques such as PCA and clustering.} have been used for the segmentation of EM images, for instance using statistical analysis of pixel neighborhoods \citep{kylberg2012segmentation}, eigenvector analysis \citep{frangakis2002segmentation}, watershed and hierarchical region merging \citep{liu2012watershed,liu2014modular}, superpixel analysis and shape modeling \citep{karabaug2019segmentation}, and random forest \citep{cao2019automatic}. However, the past few years marked a dominance of deep learning (DL) in this domain, similarly to the trends of segmentation in light microscopy and other medical imagining modalities \citep{liu2021survey,litjens2017survey}. Compared to traditional image analysis and machine learning with handcrafted features, deep learning segmentation reduces or removes the need for domain knowledge of the specific imaged sample to extract relevant features \citep{liu2021survey}.

The popularity of DL segmentation led to the development of DL plug-ins for many of the routinely used biomedical image analysis software tools like CellProfiler \citep{carpenter2006cellprofiler}, ImageJ \citep{schindelin2012fiji}, Weka \citep{carreras17}, and Ilastik \citep{berg2019ilastik}, which had previously been limited to traditional image processing methods or shallow learning.  Moreover, it led to the development of specialized tools that enable biologists to train and use DL networks with the aid of graphical user interfaces \citep{von2021,belevich2021deepmib}.

We review the recent progress of automatic image segmentation in EM, with a focus on the last six years that marked significant progress in both DL-based semantic and instance segmentation, while also giving an overview of the main DL architectures that enabled this progress.


 
The manuscript is organized as follows: Section~\ref{sec:strategy} describes the literature search strategy used for this review.~Section~\ref{sec:Section4} presents the benchmark datasets, which have been key for the progress of the segmentation methods. Section~\ref{sec:background} lays the background about the main neural network architectures for 2D and 3D segmentation of EM datasets. Sections \ref{sec:supervised} and \ref{sec:self_unsupervised} review the papers that propose new methodologies for semantic and instance segmentation with different DL approaches. These are followed by Section~\ref{sec:metrics}, which describes the evaluation metrics used in the reviewed papers. Finally, Section \ref{sec:disc} provides an outlook of the overall progress of this field along with a discussion on future prospects. 

\setcounter{footnote}{1}
\FloatBarrier
\begin{table}[h]
 \caption{Main large-scale EM techniques. More information is given in the MyScope website\protect\footnotemark and the reviews by \protect\cite{peddie2014exploring}, \protect\cite{titze2016volume} and \protect\cite{Kievits2022}. The last row shows example 2D images and 3D stacks of such technologies except STEM, an example of which is shown in Fig.~\ref{fig:Fig1}.}
 \label{tab:Table1}
\footnotesize
    \centering
\begin{adjustbox}{angle=90}
\begin{tabular}{P{6.1cm}p{11cm}} 
\toprule
\textbf{2D EM} & \textbf{Data acquisition technique}
\\
\midrule
Transmission Electron Microscopy (TEM) &  A widefield electron beam illuminates  an ultra-thin specimen and transmitted electrons are detected on the other side of the sample. The structure that is electron dense appears dark and others appear lighter depending on their (lack of) scattering. \\
\midrule
Scanning Electron Microscopy (SEM)  &  The raster scanning beam interacts with the material and can result in backscattering or the formation of secondary electrons. Their intensity reveals sample information.\\
\midrule
Scanning Transmission Electron Microscopy (STEM)  &  SEM on ultrathin sections and using a detector for the transmitted electrons.  \\
\toprule
\textbf{3D EM} & \\
\midrule
Serial section TEM (ssTEM) or SEM (ssSEM)  & Volume EM technique for examining 3D ultrastructure by scanning adjacent ultrathin (typical 60-80nm) sections using TEM or SEM, respectively.\\
\midrule
Serial Block-Face scanning EM (SBF-SEM) & The block face is scanned followed by removal of the top layer by a diamond knife (typical 20-60nm) and the newly exposed block face is scanned. This can be repeated thousands of times.  \\
\midrule
Focused Ion Beam SEM (FIB-SEM) & Block face imaging as above, but sections are repeatedly removed by a focused ion beam that has higher precision than a knife (typically down to 4nm), making it suitable for smaller \RA{samples}.   \\  
\midrule
Cryo-electron tomography (Cryo-ET) & It captures a series of 2D projection images of a flash-frozen specimen from different angles, and then uses computational reconstruction methods to generate a 3D model or tomogram.\\
\bottomrule
\end{tabular}
\end{adjustbox}
\begin{adjustbox}{angle=90}
\begin{tabular}{c@{\hspace{1mm}}c@{\hspace{1mm}}c@{\hspace{1mm}}c@{\hspace{1mm}}c@{\hspace{1mm}}c} 

            \includegraphics[width=.13\textwidth, height=.13\textwidth]{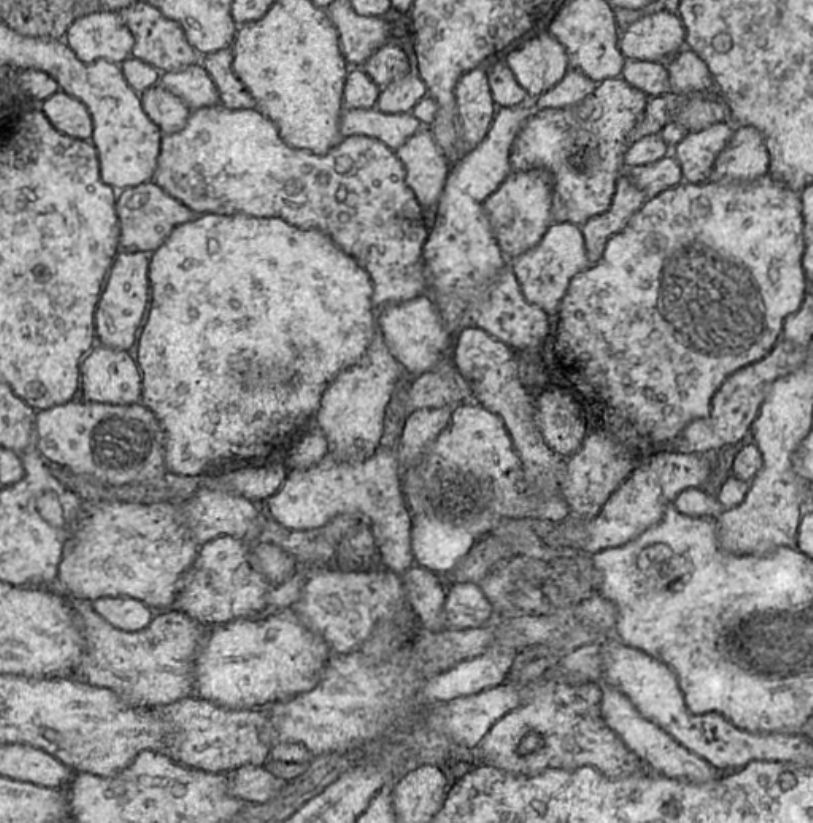}&
            \includegraphics[width=.13\textwidth, height=.13\textwidth]{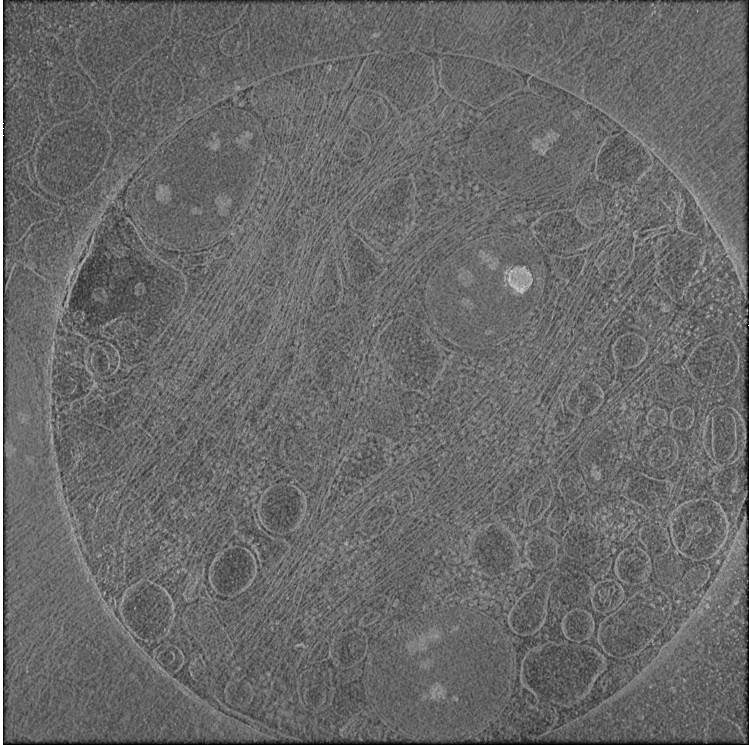}&
            \includegraphics[width=.13\textwidth, height=.13\textwidth]{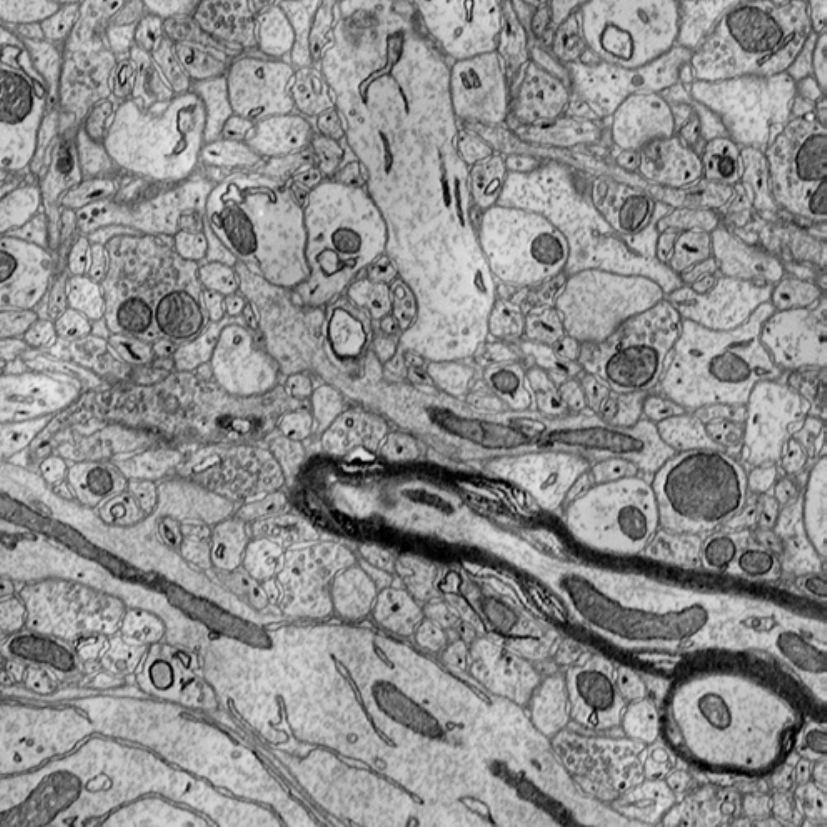}&
            \includegraphics[width=.16\textwidth, height=.14\textwidth]{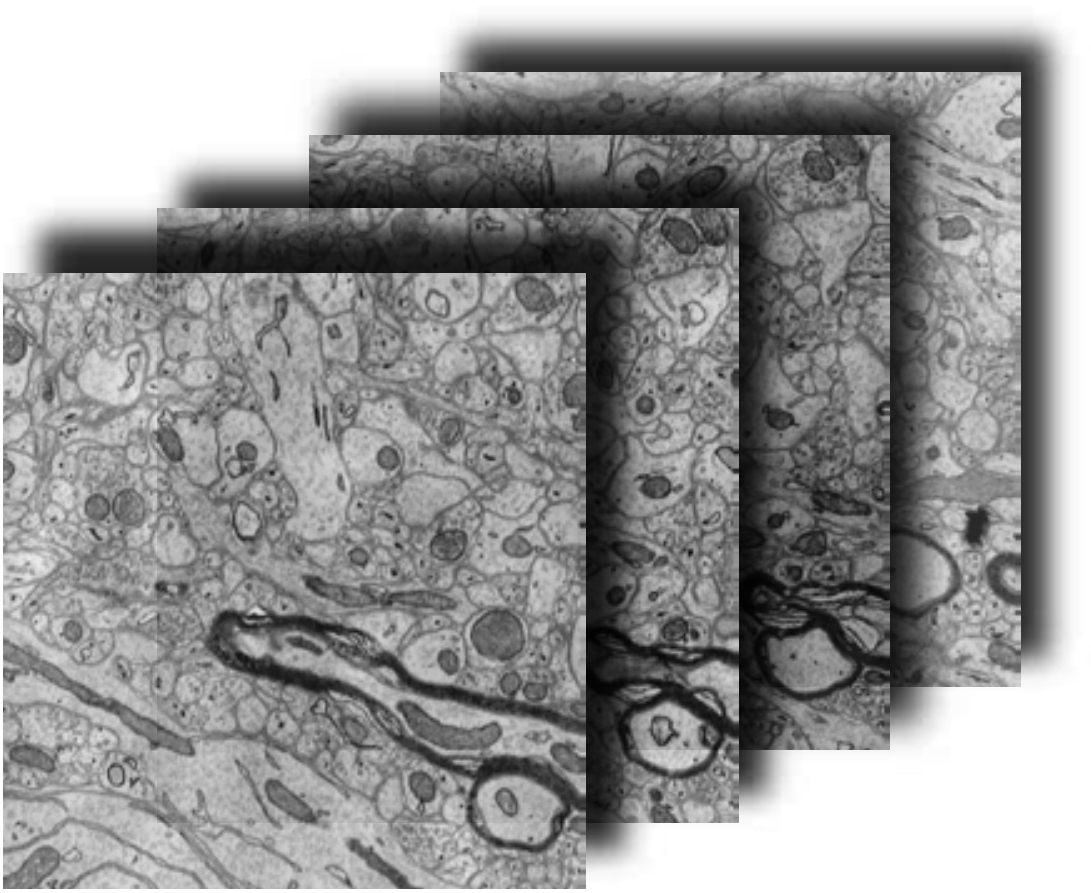}&
             \includegraphics[width=.16\textwidth, height=.13\textwidth]{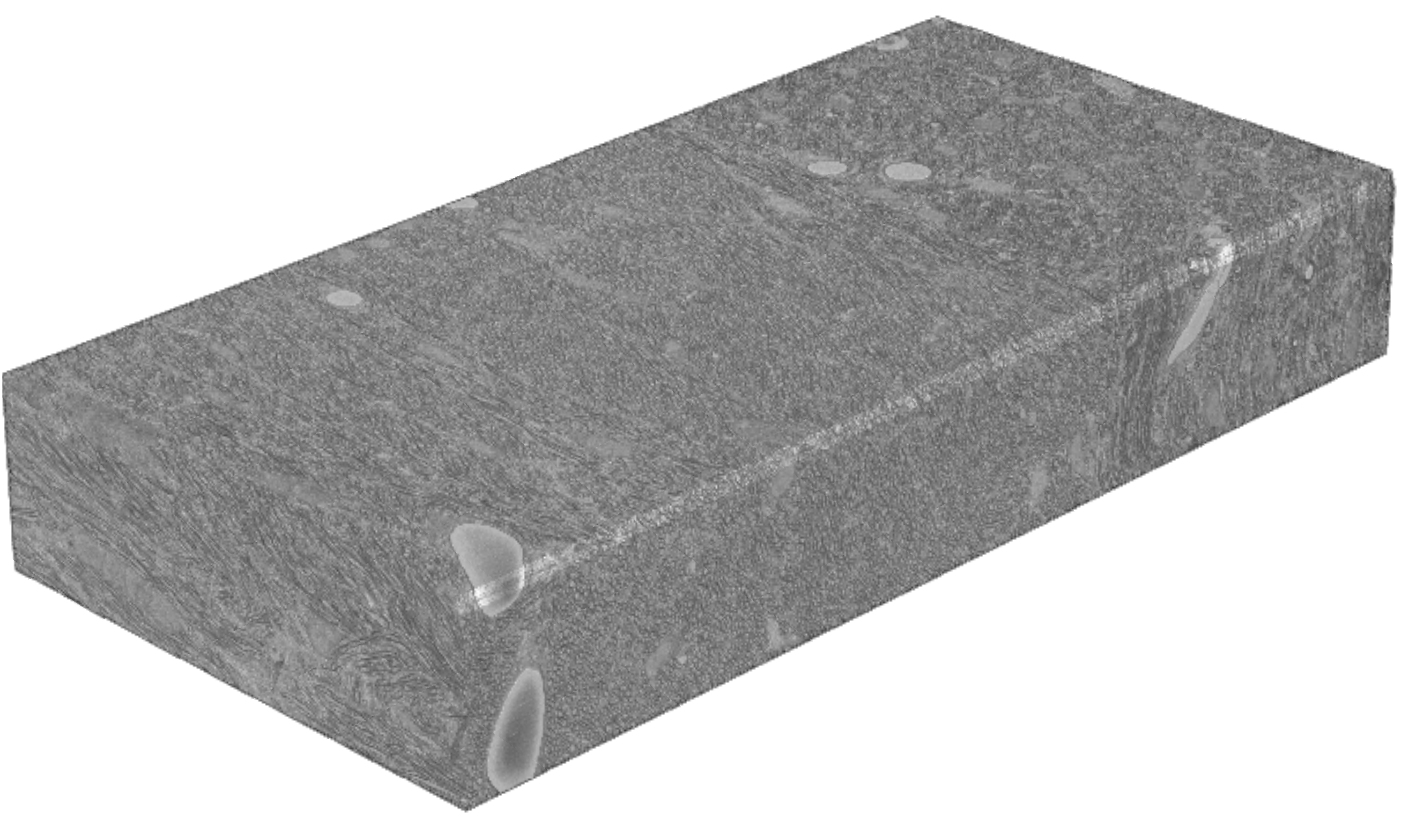}&
            \includegraphics[width=.16\textwidth, height=.16\textwidth]{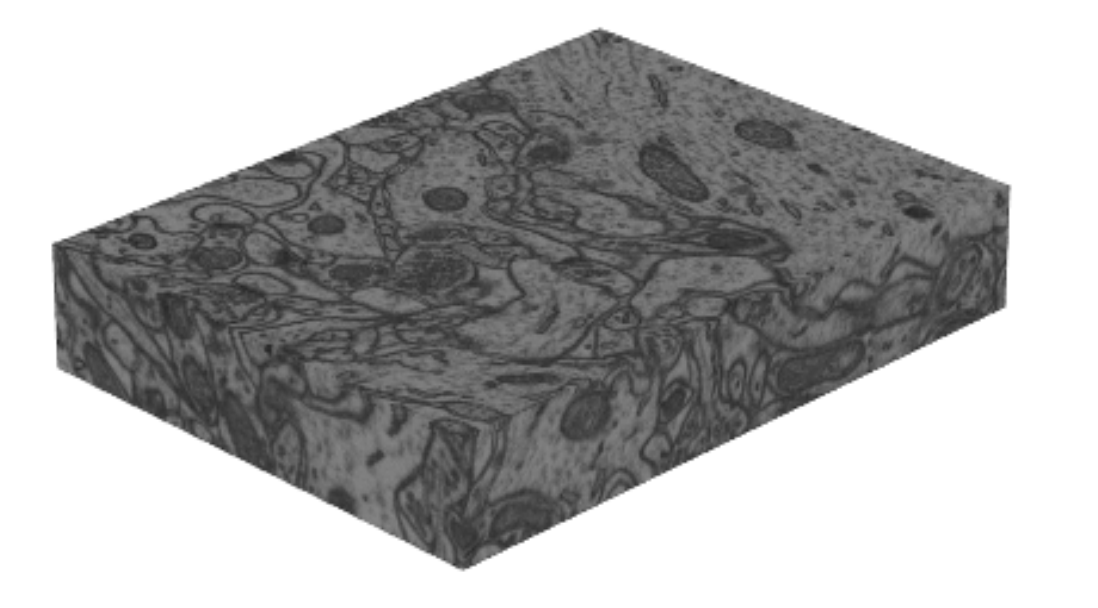}\\
            
            \textbf{TEM 2D} & \textbf{Cryo-ET 2D} & \textbf{SEM 2D} &\textbf{ssSEM volume - 2D sections} & \textbf{SBF-SEM volume} & \textbf{FIB-SEM}\\
            \cmidrule{3-4}
              \citep{ciresan2012deep}  & \citep{chen2017convolutional}  & \multicolumn{2}{c}{\citep{kasthuri2015saturated}} &\citep{abdollahzadeh2021deepacson} & \citep{lucchi2011supervoxel}\\
          
 \bottomrule

\end{tabular}
\end{adjustbox}
\end{table}
\FloatBarrier
\footnotetext{\url{https://myscope.training/}}

\section{Strategy of literature search}
\label{sec:strategy}
Our survey strategy is motivated by the following questions:
\begin{itemize}
\setlength\itemsep{-0.5em}
\item Which datasets are accessible for EM analysis, what are their challenges and what role do they play in DL research?
\item How is EM image (semantic and instance) segmentation being addressed by fully/semi/un/self-supervised DL pipelines?
\end{itemize}

To answer these questions, the following search query was used in Pubmed, Web of Science, and Google Scholar on words in titles (TI) only, restricted to 2017-2022: TI=((electron microscopy OR EM) AND (segmentation OR semantic OR instance OR supervised OR unsupervised OR self-supervised OR semi-supervised)), and title or abstracts containing (deep learning, segmentation, electron microscopy) on Google Scholar. Results from the query that were outside the scope of this study, such as deep learning in material sciences and methods based on traditional image processing (pre-DL era), were excluded. The forward and backward snowballing technique was then used to compile the final list of 38 papers. 
  
\begin{figure}[h!]
    \centering
    \includegraphics[width=0.37\textwidth,trim = {0.05cm 0 0 0},clip]{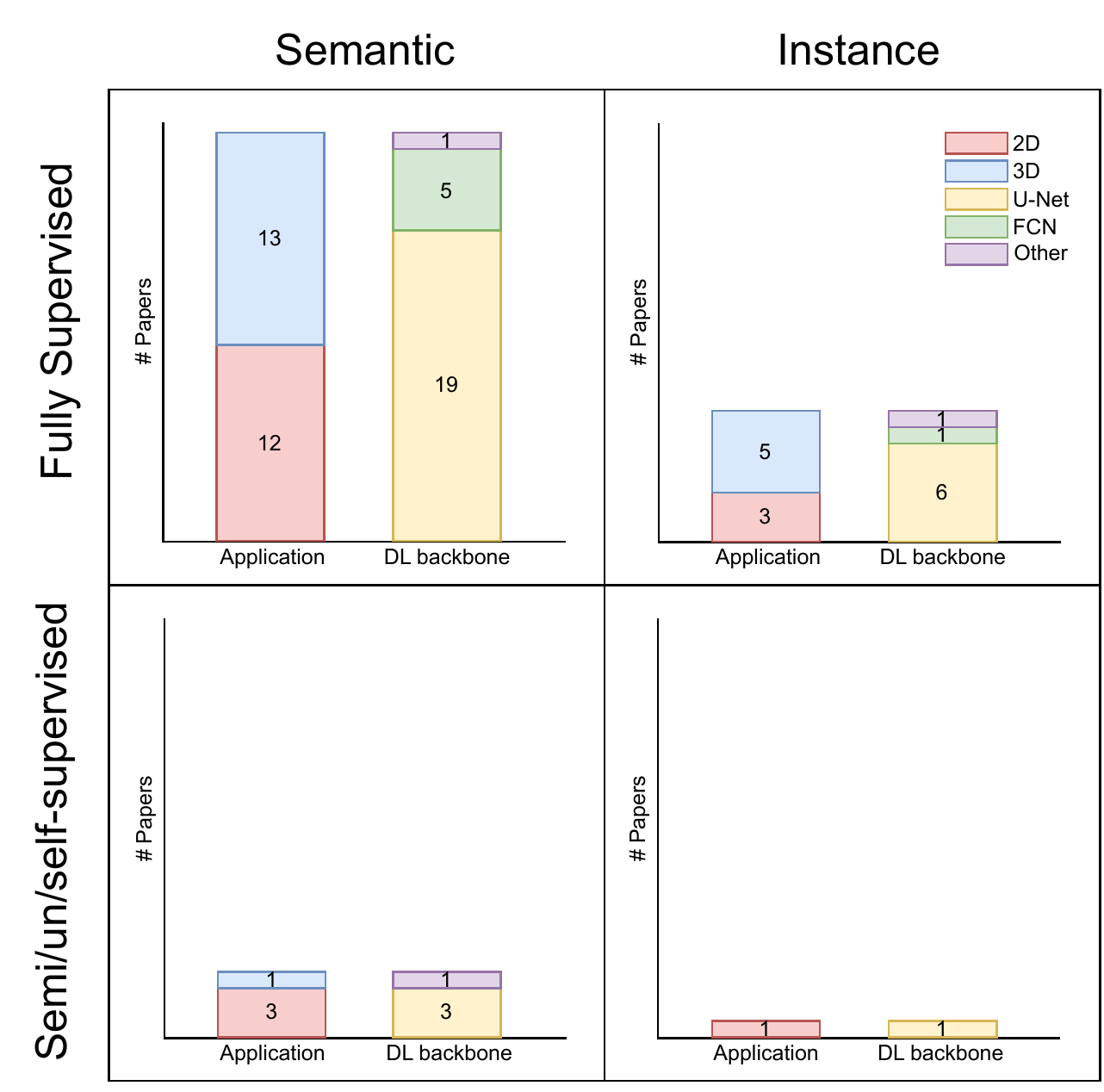}\caption{Categorization of the 38 papers reviewed in this survey. The papers are first categorized on the learning paradigm (fully vs. semi/un/self-supervised) and on the segmentation type (semantic vs. instance). Each quadrant shows the distributions of applications (2D vs. 3D) and DL backbones (U-Net vs. FCN vs. Other) of the papers that use the corresponding learning and segmentation approaches.}
    \label{fig:SearchResultSummary}
\end{figure}

Fig.~\ref{fig:SearchResultSummary} summarizes this collection of 38 papers in terms of learning technique (fully supervised or not), segmentation type (semantic or instance), application (2D or 3D) and the underlying modeling backbone. Before reviewing these papers, we discuss the key EM datasets and describe the evolution of DL architectures, which are two crucial components that have been permitting the progress of EM segmentation analysis.

\RA{\section{Collections of key EM datasets}
\label{sec:Section4}
Collections of labeled and unlabeled EM images have played a significant role in advancing DL research for EM segmentation, and some were associated with notable segmentation competitions and challenges.~This section provides the details of all collections used by the 38 papers in this survey.~Table~\ref{tab:datasets} reports the main properties of these datasets and below is an in-depth discussion of their characteristics and the challenges they address.~The discussion is categorized according to the EM modality used to acquire the datasets.

\subsection{Serial section TEM and SEM datasets}
Serial-section transmission or scanning EM (ssTEM or ssSEM) is used for studying synaptic junctions and highly-resolved membranes in neural tissues. Advances in microscopy techniques in serial section EM have enabled the study of neurons with increased connectivity in complex mammalian tissues (such as mice and humans) and even whole brain tissues of smaller animal models, like the fruit fly and zebrafish. This imaging approach visualizes the generated volumes in a highly anisotropic manner, i.e. the $x$- and $y$-directions have a high resolution, however, the $z$-direction has a lower resolution, as it is reliant on serial cutting precision.

The Drosophila larve dataset (\#1)\footnote{$\#n$ refers to the entry $n$ in Table~2.} of the ISBI 2012 challenge was the first notable EM dataset for automatic neuronal segmentation, featuring two volumes with 30 sections each. The main challenge of that dataset is to develop algorithms that can accurately segment the neural structures present in the EM images. The success of deep neural networks as pixel classifiers in the ISBI 2012 challenge \citep{ciresan2012deep} paved the way for deep learning in serial section EM segmentation. Recently, a connectome of an entire brain of a Drosophila fruit fly has been published by \citet{Winding2023}, and will serve as a new resource for various follow-up works.

The CREMI3D dataset (\#2) consists of three large and diverse sub-volumes of neural tissue along with ground truth annotations for training and evaluation purposes, and was part of a competition at the MICCAI 2016 conference.~The dataset comes from a full adult fly brain (FAFB) volume and contains 213 teravoxels. It was imaged at the synaptic resolution to understand the functioning of brain circuits (connectomics)  and its goal was to segment neurons, synapses, and their pre-post synaptic partners. The CREMI3D dataset is part of the FlyEM project and since its inception, it has been used to evaluate various image analysis methods for neural circuit reconstruction, including DL approaches such as convolutional neural networks (CNNs) and recurrent neural networks (RNNs). 

The SNEMI3D dataset (\#3) consists of a volume of 100 ssSEM images of the neural tissue from a mouse cortex.~It is a subset of the largest mouse neocortex dataset imaged by \citet{kasthuri2015saturated} using an automated ssSEM technique and hence is also known as the Kasthuri dataset. The dataset was created as part of the ISBI 2013 challenge on segmentation of neural structures in EM images. The main challenge of this dataset is to develop algorithms that can accurately segment the neuronal membranes present in the EM images and reconstruct a 3D model of the tissue. This is a difficult task due to the large size of the dataset and the complexity of the neural structures, namely axons, dendrites, synapses, and glial cells. 

The Kasthuri++ and Lucchi++ (\#4, \#9) datasets were introduced by \citet{casser2018fast} with corrected annotations of Kasthuri and Lucchi. The Kasthuri dataset, which is used for dense reconstructions of the neuronal cells was corrected for the jaggedness between inter-slice components as they were not accurate. The Lucchi dataset is a FIB-SEM dataset used for the segmentation of mitochondria in the mouse neocortex. It was corrected for consistency of all annotations related to mitochondrial membranes, as well as to rectify any categorization errors in the ground truth. 

\FloatBarrier
\begin{table*}[h]
    \scriptsize
    \centering
    \captionsetup{width=1.3\textwidth}
    \caption{Key datasets from studies that perform high-resolution automated (volume) EM segmentation using deep learning. The abbreviations of the (sub) cellular structures are defined in the legend.} 
    \label{tab:datasets}
\hspace*{-1.1in}\begin{threeparttable}[t]
\begin{tabular}{P{0.05cm}P{2cm}P{1.5cm}P{2.3cm}P{1.6cm}P{2.5cm}P{3.2cm}P{4cm}}
    \toprule
   \textbf{\#}& \textbf{Dataset} & \textbf{Acquisition} & \textbf{Region}	&\textbf{Pixel/Voxel size} ($nm$) & \textbf{Pixels} &\textbf{Labeled (sub) cellular structures}& \textbf{Public repository}\\    
       \midrule
        1&ISBI 2012/ Drosophila VNC   & ssTEM &Nervous cord (Drosophila) &   $4 \times 4 \times 50$ & $512 \times 512 \times 30$&NM &  \url{https://imagej.net/events/isbi-2012-segmentation-challenge} \\
        \midrule      
        2&MICCAI 2016/ CREMI3D  & ssTEM &Adult fly brain (Drosophila) & $4 \times 4 \times 40$& $1250 \times 1250 \times 125$&NM, S, SP  &\url{https://cremi.org}\\
        \midrule
        3&ISBI 2013/ SNEMI3D / Kasthuri &ssSEM& Neocortex (Mouse)&   $3 \times 3 \times 30$ & 
        $1024 \times 1024 \times 100$& NM &   \url{https://snemi3d.grand-challenge.org/}\\ 
        \midrule
        4&Kasthuri++ &ssSEM& Neocortex (Mouse)&   $3 \times 3 \times 30$ &$1643\times 1613 \times 85$&  M, NM & \url{ https://casser.io/connectomics}\\
        \midrule 
         5&Xiao& ssSEM &Cortex (Rat)& $2 \times 2 \times 50$&$8624 \times 8416 \times 20$& M &\url{http://95.163.198.142/MiRA/mitochondria31/}\\
        \midrule
        6& MitoEM & ssSEM & Cortex(Human, rat) &$8 \times 8 \times 30$&$4096\times4096 \times 1000$&M &  \url{https://mitoem.grand-challenge.org/} \\
        \midrule
        7&NucMM  & ssSEM & Whole brain (Zebrafish) & $4 \times 4 \times 30$& $1450 \times 2000 \times 397$&N &\url{https://nucmm.grand-challenge.org/} \\
        \midrule
        8&Lucchi / EPFL Hippocampus & FIB-SEM & Hippocampus (Mouse) &  $5 \times 5\times 5$ & $1024 \times 768 \times 165$ &M&\url{https://www.epfl.ch/labs/cvlab/data/data-em/}\\
        \midrule
        9 &  Lucchi++ & FIB-SEM & Hippocampus (Mouse) &  $5 \times 5\times 5$ & $1024 \times 768 \times 165$ & M&\url{https://casser.io/connectomics}\\
        \midrule
        10&FIB-25  & FIB-SEM &Optic lobe (Drosophila)&  $8 \times 8 \times 8$ & $520 \times 520 \times 520$ &N, S& \url{http://research.janelia.org/FIB-25/FIB-25.tar.bz2}\\
        \midrule
        11&OpenOrganelle & FIB-SEM & Interphase HeLa, Macrophage, T-cells& $8 \times 8\times 8$ & Varying sizes &CN, CH, EN, ER, ERN, ERES, G, LP, L, MT, NE, NP, Nu, N, PM, R, V&\url{https://openorganelle.janelia.org}\\
        \midrule
        12 & Cardiac mitochondria & FIB-SEM& Heart muscle (Mouse) &$15 \times 15 \times 15$ &$1728 \times 2022 \times 100$& M &  \url{http://labalaban.nhlbi.nih.gov/files/SuppDataset.tif}\\
        \midrule
        13&   UroCell  & FIB-SEM & Urothelial cells (Mouse) & $16 \times 16 \times 15$&  5 subvolumes of $256 \times 256 \times 256$& G, L, M, V &  \url{https://github.com/MancaZerovnikMekuc/UroCell}\\
      \midrule
       14&Perez  & SBF-SEM & Brain (Mouse) & $7.8 \times 7.8 \times 30$& $16000 \times 12000 \times 1283$ &L, M, Nu, N & \url{https://www.sci.utah.edu/releases/chm_v2.1.367/}  \\
      \midrule
       15&SegEM & SBF-SEM & Mouse cortex  & $11\times11\times26$ &279 volumes of $100 \times 100 \times 100$ &NM&\url{https://segem.rzg.mpg.de/webdav/SegEM_challenge/}\\
     \midrule
     16&Guay& SBF-SEM& Platelets (Human)&$ 10 \times 10 \times 50$&$800 \times 800 \times 50$&Cell, CC, CP, GN, M&\url{https://leapmanlab.github.io/dense-cell/}\\
\midrule

    17&  Axon&SBF-SEM & White matter (Mouse) &   $50 \times 50\times 50 $  & $1000 \times 1000 \times 3250 $ & A, M, My, N & \url{http://segem.brain.mpg.de/challenge/}\\
    \midrule
   18&CDeep3M-S & SBF-SEM & Brain (Mouse) & $2.4 \times 2.4 \times 24$& $16000 \times 10000 \times 400$& M, NM, Nu, V & \url{https://github.com/CRBS/cdeep3m} \\ 
    \midrule
    19& EMPIAR-10094& SBF-SEM& HeLa cells& $10 \times 10 \times 50$& $8192 \times 8192 \times 517$& Unlabeled & \url{http://dx.doi.org/10.6019/EMPIAR-10094} \\
    \midrule
    20 & CEM500K & All of the above & 20 regions (10 organisms)& $2\times2\times2$ to $20\times20\times20$ & $224 \times 224 \times 496544$& Unlabeled& \url{https://www.ebi.ac.uk/empiar/EMPIAR-10592/} \\
      \midrule
      21 & CDeep3M-C & Cryo-ET & Brain (Mouse) & $1.6 \times 1.6 \times 1.6$ & $938 \times 938 \times 938$& NM, V & \url{https://github.com/CRBS/cdeep3m} \\
      \midrule
      22 & Cellular Cryo-ET& Cryo-ET&PC12 cells & $2.8 \times 2.8 \times 2.8$ & $938 \times 938 \times 938$ & L, M, PM, V & \url{https://www.ebi.ac.uk/emdb/EMD-8594}\\
 \bottomrule
\end{tabular}
 \end{threeparttable}
 \justifying
 \hspace*{-1.1in}\begin{minipage}{20cm}
A - Axons, CC - Canalicular channel, CH - Chromatin, CN - Centrosome, CP - Cytoplasm, D - Dendrites, EN - Endoplasmic Reticulum, ERES - Endoplasmic Reticulum Exit Site, G - Golgi, GC - Glial cells, GN - Granules, L - Lysosome, LD - Lipid Droplet, M - Mitochondria, MT - Microtubule, My - Myelin. N - Nucleus, NE - Nuclear Envelope, NM - Neuronal membrane, NP - Nuclear Pore, Nu - Nucleolus, PM - Plasma Membrane, R - Ribosome, S -Synapse, SP - Synaptic partners, V - Vesicle.
\end{minipage}
\end{table*}
\FloatBarrier

The Xiao (\#5) dataset for mitochondria segmentation was collected from a rat brain by \citet{xiao2018automatic} using advanced ssSEM technology. Automated cutting was used to produce 31 sections, each with an approximate thickness of 50 nm for segmenting mitochondria. The ground truth dataset was prepared through 2D manual annotation and image registration of serial-section images, which was made publicly available for accelerating neuroscience analysis. 


Mito-EM (\#6) \citep{wei2020mitoem} introduced the largest mammalian mitochondria dataset from humans (MitoEM-H) and adult rats (MitoEM-R). It is about 3600 times larger than the standard dataset for mitochondria segmentation (Lucchi) containing mitochondria instances of at least 2000 voxels in size. Complex morphology such as mitochondria on a string (MOAS) connected by thin microtubules or instances
entangled in 3D were captured using ssSEM. The MitoEM dataset was created to provide a comprehensive view of the ultrastructure of mitochondria and to facilitate a comparative study of mitochondrial morphology and function in rats and humans.

The NucMM dataset (\#7) \citep{lin2021nucmm} contains two fully annotated volumes; one that contains almost a whole zebrafish brain with around 170,000 nuclei imaged using ssTEM; and another that contains part of a mouse visual cortex with about 7,000 nuclei imaged using micro-CT. Micro-CT or micro-computed tomography uses X-rays to produce 3D images of objects at low resolution and hence is not a part of this review. The large-scale nuclei instance segmentation dataset from ssTEM covers 0.14${mm}^3$ of the entire volume of the zebrafish brain at $4 \times 4 \times30 $ nm/voxel. As most of the nuclei segmentation datasets are from light microscopy at the $\mu m$ scale, the dataset was downsampled to $512 \times 512 \times 480 $ nm/voxel. 

\subsection{FIB-SEM datasets}
FIB-SEM generates datasets that are ideal for automated connectome tracing and for examining brain tissue at resolutions lower than $10 \times 10 \times 10$ nm. The method can produce sections with a thickness of $4$ nm, but the volumes are typically smaller in comparison to other techniques, due to their high $z$-resolutions. 

The Lucchi dataset (\#8) is an isotropic FIB-SEM volume imaged from the hippocampus of a mouse brain, and it has the same spatial resolution along all three axes. This dataset has now become the de facto standard for evaluating mitochondria segmentation performance. 
Efforts to expand FIB-SEM to larger volumes were made by \citet{takemura2015synaptic} who compiled the FIB-25 (\#10) dataset by reconstructing the synaptic circuits of seven columns in the eye region of a Drosophila's brain. FIB-25 contains over 10,000 annotated neurons, including their synaptic connections, and is one of the most comprehensive EM datasets of the Drosophila brain to date. It was created to provide a detailed map of the neural circuits in the Drosophila brain and to facilitate the study of neural connectivity and information processing. The dataset is publicly available and can be accessed through the FlyEM project website. 

Enhanced FIB-SEM techniques have also enabled high-throughput and reliable long-term imaging for large-scale EM ($10^3$ to $3 \times 10^7 \mu m^3$), such as the OpenOrganelle atlas (\#11) of 3D whole cells and tissues of \citet{xu2021open}. The datasets for the 3D reconstruction of cells were made open-source under the OpenOrganelle repository for exploring local cellular interactions and their intricate arrangements.

Other FIB-SEM datasets include ones requiring a high-resolution analysis of 3D organelles in important tissues of the heart muscle and urinary bladder. Cardiac mitochondria (\#12) is a FIB-SEM dataset introduced to segment mitochondria in cardiomyocytes \citep{khadangi2021stellar}. The FIB-SEM technique was needed to better characterize diffusion channels in mitochondria-rich muscle fibers. Isotropic voxels at 15 nm resolution were imaged according to the set of experiments performed by \citet{glancy2015mitochondrial}. The UroCell (\#13) from FIB-SEM was imaged by \citet{mekuvc2020automatic} to focus on mitochondria and endolysosomes and was further extended to Golgi apparatus and fusiform vesicles. The dataset is unique as it is publicly available for further analysis of the epithelium cells of the urinary bladder, where the organelles form an important component in maintaining the barrier between the membrane of the bladder and the surrounding blood tissues. 

\subsection{SBF-SEM datasets}

Connectomics research was also based on popular datasets imaged using SBF-SEM \citep{ helmstaedter2013connectomic, briggman2011wiring}. Imaging using SBF-SEM produces anisotropic sections but does not need image registration and avoids missing sections in comparison to serial-sectioning TEM/SEM, as the technique images the sample intact on a block surface.  Such a technique also enabled imaging large volumes for studying the organization of neural circuits and cells across hundreds of microns through millimeters of neurons in a $z$-stack. 

The Perez dataset (\#14) \citep{perez2014workflow} involved the acquisition of 1283 serial images from the hypothalamus's suprachiasmatic nucleus (SCN), a small part of the mouse brain, to produce an image stack with tissue dimensions approximately measuring 450,000 $\mu m^3$. The large acquired volume was downsampled from 3.8 to 7.8 nm/pixel in the $x-y$ resolution to scale up the processing of these tetra-voxel-sized SBF-SEM images. It was introduced for the automatic segmentation of mitochondria, lysosomes, nuclei, and nucleoli in brain tissues. 

SegEM (\#15) introduced an EM dataset acquired using SBF-SEM from the mouse somatosensory cortex \citep{berning2015segem}. The images in the SegEM dataset are provided with corresponding segmentation labels for dendrites, axons, and synapses. The labels were generated using a semi-automated approach which involved a combination of skeleton annotations and machine learning algorithms to trace long neurites accurately. Since then, SegEM has been used for benchmarking popular models like flood-filling networks that test the efficiency of algorithms on volume-spanning neurites.

The Guay dataset (\#16) is a fully annotated dataset of platelet cells from two human subjects and was designed for dense cellular segmentation \citep{guay2021dense}. It has also been used for large-volume cell reconstruction along with mitochondria, nuclei, lysosomes, and various granules inside the cells. 


The Axon dataset (\#17) is a collection of SBF-SEM images of white matter tissue from rats, captured at a lower resolution of 50 nm/pixel \citep{abdollahzadeh2021deepacson}. The low-resolution image stack of 130000 $\mu m^3$ was enough to resolve structures like myelin, myelinated axons, mitochondria,
and cell nuclei. A wide field of view employing low-resolution SBF-SEM stacks was considered important for quantifying metrics such as myelinated axon tortuosity, inter-mitochondrial distance, and cell density.

CDeep3M proposed two new datasets from SBF-SEM and cryo electron tomography (cryo-ET) for automatic segmentation. The first one, CDeep3M-S (\#18), is a large SBF-SEM dataset for membrane, mitochondria, and synapse identification from the cerebellum and lateral habenula of mice. Imaged at $2.4$ nm pixel size, a cloud implementation of the latest architecture for anisotropic datasets was used to segment structures such as the neuronal membrane, synaptic vesicles mitochondria, and nucleus in brain tissues. The second dataset, CDeep3M-C (\#21), was from cryo-ET and is explained further in subsection \ref{cryo_dataset}. 

The EMPIAR-10094 dataset (\#19) consists of EM images of cervical cancer ``HeLa" cells imaged using SBF-SEM. The dataset is imaged at $8192 \times 8192$ pixels over a total of 518 slices, and consists of different HeLa cells distributed in the background of the embedding resin. The dataset has been made publicly available with no labels and has mostly been used for delineating structures such as plasma membranes and nuclear envelopes.

Unlabeled datasets, such as CEM500K, from various unrelated experiments and EM modalities for solving the segmentation of a particular structure seem promising. The CEM500k (\#20) is an EM unlabeled dataset containing around 500,000 images from various unrelated experiments and different EM modalities for cellular EM. The images from different experiments were standardized to 2D images of size $512 \times 512$ pixels with pixel resolutions ranging from 2 nm in datasets from serial section EM and ${\sim}20$ nm for SBF-SEM. The dataset was further filtered by removing duplicates and low-quality images in order to provide robustness to changes in image contrast and making it suitable for training modeling techniques.

\subsection{Cryo-ET datasets}
\label{cryo_dataset}
Electron tomography (ET) is used to obtain 3D structures of EM sections using the tilt-series acquisition technique. Cryo-ET does so at cryogenic temperatures to image vitrified biological samples. Attempts for segmentation on cryo-ET can be found by \citet{MOUSSAVI2010134} and in the review of \citet{carvalho20183d}. The identification of macromoleular structures is beyond the scope of this review.
Cryo-ET presents challenges in visualizing and interpreting tomographic datasets due to two main factors. Firstly, sample thickness increases as the tilt angle increases, leading to an artifact known as the ``missing wedge" and reduced resolution in the $z$-direction. Secondly, vitrified biological samples are sensitive to electron dose, resulting in a low signal-to-noise ratio and difficulties in distinguishing features of interest from background noise. As the resolution capacity of TEM decreases with the increase in sample thickness, focused ion beam (FIB) milling can be used to obtain a high-resolution tomogram. Cryo-FIB SEM is an evolving technology for cellular imaging that is rapidly being used in recent years. This is mainly attributable to its ability to image larger specimens that may be too thick for cryo-ET, such as whole cells or tissues.


CDeep3M-C (\#21) is a cryo-ET dataset for the segmentation of vesicles and membranes from the mouse brain \citep{haberl18}. At a voxel size of 1.6 nm, it was used to digitally recreate a tiny section (approximately $1.5 \times 1.5 \times 1.5$$ \mu m^3$) of a high-pressure frozen tissue. The final volume was built from 7 sequential tomograms (serial sections), each created by tilting a sample every $0.5^\circ$ in an electron beam from $-60^\circ to +60^\circ$.  The cellular cryo-ET dataset (\#22) was acquired at low magnification for annotation and qualitative cellular analysis of organelles like mitochondria, vesicles, microtubules, and plasma membrane \citep{chen2017convolutional}. The PC12 cell line was reconstructed using 30 serial sections imaged at $850 \times 850 \times 81$ pixel size at 2.8 nm resolution. The tomograms of platelets and cyanobacteria utilized in that work are from previously published datasets \citep{wang2015electron, dai2013visualizing}.
}
\section{Background of backbone deep learning networks for EM semantic and instance segmentation}\label{sec:background}The rapid progress of DL methods, in particular CNNs, has had a great impact on advancing segmentation of EM images, as well as other medical images of various modalities \citep{litjens2017survey,shen2017deep}, including light microscopy \citep{xing2017deep,liu2021survey}. Deep learning in EM analysis has also been addressed in the reviews by \citet{treder2022applications} and \citet{ede2021deep}. The former gives a broad overview of different EM applications in both physical and life sciences and the latter provides a practitioner's perspective focused on the hardware and software packages to perform DL-based EM analysis.~In contrast, this review provides an in-depth view of fully/semi/self/un-supervised deep learning methods for the semantic and instance segmentation in (sub)cellular EM. This section covers the main milestones in the progression of network architectures and their key attributes, which are necessary to put in context the 38 papers that are reviewed in this work.



Semantic segmentation in EM images is the identification of objects or subcellular organelles in such a way that each pixel is mapped to a specific class. This is different than instance segmentation, which refers to the process of dividing an image into multiple segments, each corresponding to a unique object or instance. Instance segmentation is particularly important in the study of cellular structures and their interactions, as it allows for the identification and quantification of individual objects in large-scale datasets.

\begin{figure*}[t]
        \includegraphics[width=1.02\textwidth, height=0.45\textwidth]{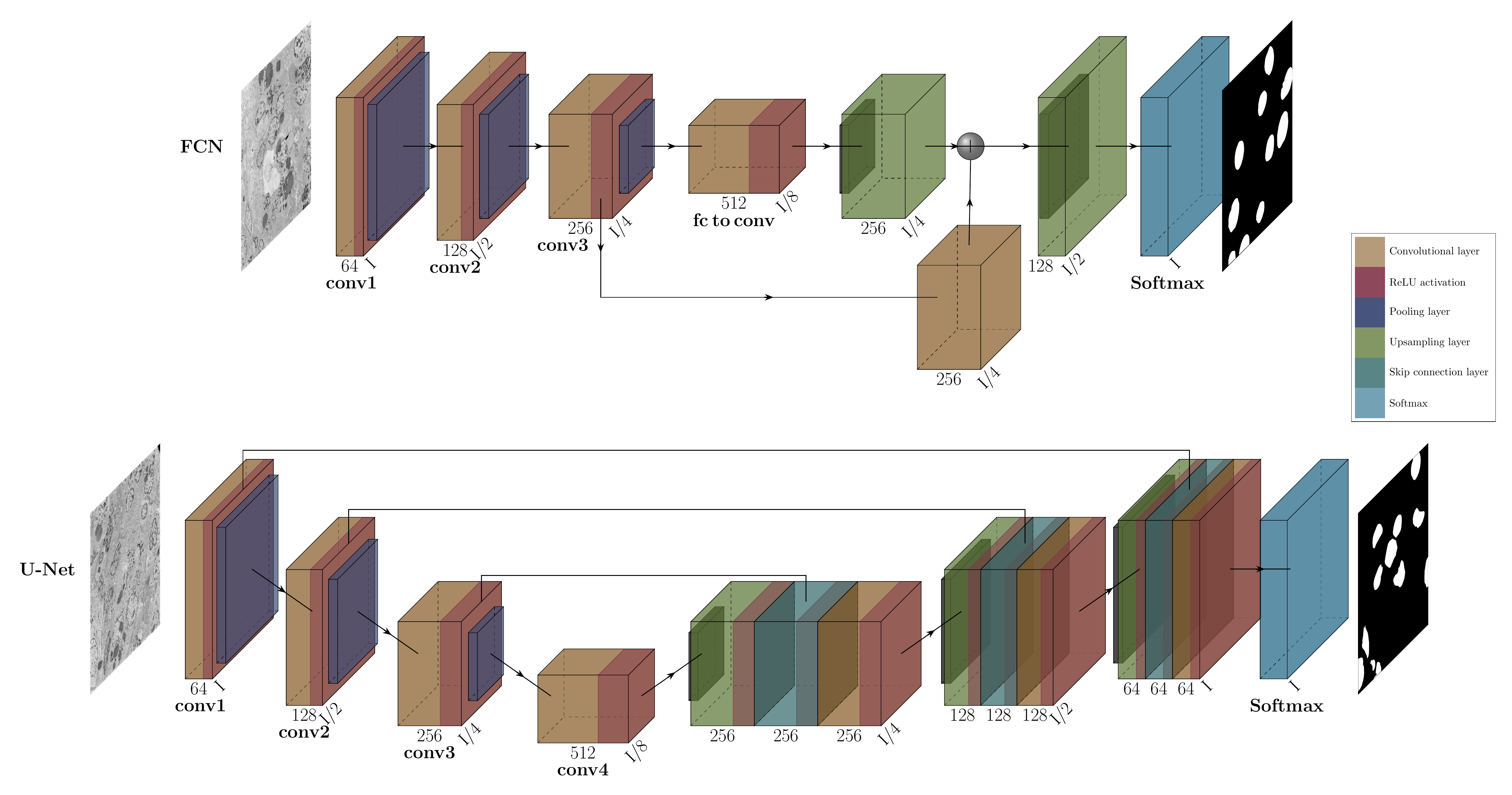} \caption{Encoder-decoder networks of the original works on FCN \citep{long2015fully} and U-Net \citep{ronneberger2015u}. Each of the fully connected (fc) layers in FCN and convolutional layers in U-Net are followed by the nonlinear activation function ReLU and max pooling. In FCN, the fully connected layers are then converted to convolutional layers via the `fc to conv' component. The last layer uses a softmax function to assign a probability class score to each pixel. The FCN decoder includes an upsampling component that is linearly combined with the low-level feature maps in the third convolutional layer of the encoder. The sizes of these feature maps are four times less than the size of the input image $I$ (denoted by $I/4$). Finally, there is a direct upsampling from $I/4$ to the original size of $I$ followed by softmax for classification. The symmetrical U-Net architecture shares the features maps in the encoder with the decoder path together with skip connections.}
    \label{fig:FCN}
\end{figure*}


The CNN designed by \citet{ciresan2012deep}, for instance, was used for the semantic segmentation of neuronal membranes in stacks of EM images. The images were segmented by predicting the label of each local region or patch covered by a convolutional filter in a sliding window approach and introduced max-pooling layers instead of sub-sampling layers. \RB{As indicated by \citet{arganda2015crowdsourcing}, it led to winning the ISBI 2012 neuronal segmentation challenge}\footnote{\url{https://imagej.net/events/isbi-2012-segmentation-challenge}}.
 
\RB{Despite its success, the method suffered from two major limitations - firstly, the sliding window approach was slow due to the redundancy of processing large overlaps between adjacent patches, and secondly, there was a trade-off between the size of the patches (context) and localization accuracy. Since the network's depth is an important factor for a larger receptive field (the size of the viewing field from which the network receives information), larger patches require deeper networks. Localization ability, however, decreases with deeper networks due to downsampling by the many max pooling layers and the use of smaller patches allows the network to see only a little context. 


Improvements in the semantic segmentation of EM images continued with the development of the Fully Convolutional Network (FCN) \citep{long2015fully} and the U-Net architecture \citep{ronneberger2015u}, Fig.~\ref{fig:FCN}. The concept of expanding a CNN to handle inputs of any size using fully convolutional layers instead of fully connected ones helped evolve dense predictions for segmentation. A skip architecture was introduced to make use of a feature spectrum that merges deep, coarse, semantic information with shallow, fine, appearance information. ~The U-Net extended an FCN network with a U-shaped topology to optimize the tradeoff between localization and context. The contracting path (encoder) captures a larger context using the downsampled features and the expanding path (decoder) upsamples features to their original size with the same number of layers making it a symmetric or U-shaped network. The skip connections between the encoder-decoder layers bypass some of the neural network layers and as a result, an alternative and shorter path is provided for backpropagating the error of the loss function, which contributed to avoiding the vanishing gradient problem \citep{krizhevsky2012imagenet}. Increased connectivity in the upsampling path within FCNs and the consideration of multi-level contexts were key to improving semantic segmentation \citep{badrinarayanan2017segnet, drozdzal2016importance}.}

DeepLab is another family of semantic segmentation networks, which have the ability to achieve robustness for different scales without increasing computational complexity \citep{chen2014semantic,chen2017deeplab,chen2017rethinking,chen2018encoder}. DeepLab architectures are based on FCNs but extended with the use of dilated (or atrous) convolutions, which were originally proposed by \cite{Fisher2016}, and image-level features. The atrous dilations are used within Atrous Spatial Pyramid Modules (ASPP), which perform multi-scale feature extraction by using multiple atrous convolutions with different dilation rates. As a backbone network, the latest DeepLab architecture, namely DeepLab v3+, uses the Residual Neural Network (ResNet) to produce image-level feature maps. The module performs parallel convolution on the feature map obtained from the ResNet backbone and outputs multiple feature maps, which are then concatenated and fed into the next layer. This allows the network to capture features of multiple scales, which is crucial for tasks like semantic segmentation. ResNet is notable for its ability to overcome the vanishing gradient problem and the degradation issue, simultaneously \citep{he2016deep}. This breakthrough was attributable to the introduction  of residual connections, which allow the network to learn residual functions, or the difference between the desired output and the current output, rather than the full function. This helps the network to learn more effectively and avoid overfitting.

\RA{The 3D segmentation of neuronal stacks was set as a challenge
in ISBI 2013. The major challenges in analyzing volume EM datasets are misalignments or missing sections due to serial sectioning, and volume anisotropy due to different resolutions in different directions. Specifically, it refers to the situation where the resolution in the $z$-axis (the depth dimension) is lower than the resolution in the $x-y$ plane.

There are three typical approaches for the analysis of 3D volumes. The first involves 2D segmentation of each image in the stack, followed by 3D reconstructions based on clustering techniques, that may range from basic watershed to complex graph cuts algorithms. The second approach is based on 3D CNNs, which can learn representations of volumetric data that include 3D spatial context. One example of such 3D CNNs is the 3D U-Net by \citet{cciccek20163d}, which was inspired by the original U-Net that uses local and larger contextual information. It was then extended into the V-Net model by \citet{milletari2016v} by adding residual stages. The HighRes3DNet is another 3D CNN based on the FCN architecture, with dilated and residual convolutions, and has been successful in obtaining accurate segmentations of neuronal mitochondria \citep{li2017compactness}. In terms of performance, both HighRes3DNet and V-Net have achieved state-of-the-art results on several medical image segmentation benchmarks. However, HighRes3DNet has been shown to have better performance on tasks involving high-resolution and multi-modal medical images, while V-Net has been shown to be more efficient in terms of computational resources and memory usage. A variant of the 3D network is the hybrid 2D-3D methodology as proposed by \citet{lee2015recursive} for the segmentation of anisotropic volumes. They utilize only 2D convolutions in the initial layers that downsample the input feature maps with high $x-y$ resolution (independent of the $z$-axis) until they are roughly isotropic to be efficiently processed by 3D convolutions. 

Graph analysis is the third approach for 3D segmentation. Graph-based methods typically involve partitioning a graph into regions or clusters based on properties such as color or intensity values, edge strength, or other image features such as shape. These methods often use graph theory algorithms, like graph cuts or minimum spanning trees, to identify regions that are distinct from one another. This may be coupled with structure-based analysis that uses certain geometrical properties to identify boundaries between objects. Global shape descriptors were used to learn the connectivity of 3D super voxels by \citet{lucchi2013learning} for segmentation using graph-cuts, addressing issues with local statistics and distracting membranes. \citet{turaga2010convolutional} suggested how CNNs can be used for directly predicting 3D graph affinities based on a structured loss function for neuronal boundary segmentation. The proposed loss function assigned scores to the edges between adjacent pixels based on their likelihood of belonging to same or different regions and also penalized their assignment for achieving incorrect predictions that violate the underlying structure of the image.

Instance segmentation involves classifying each pixel/voxel of a given image/volume to a particular class along with assigning a unique identity to pixels/voxels of individual objects. Instance segmentation using deep learning can be divided into proposal-based (top-down) and proposal-free (bottom-up) approaches. Proposal-based approaches such as RCNN, FastRCNN, and FasterRCNN are two-stage detection networks that use a deep neural network for feature extraction (encoder) and region proposals for the segmentation of objects of interest, followed by bounding box regression and classification to obtain instance segmentation \citep{liu2020deep}. Mask-RCNN \citep{he2017mask} is a popular choice for generic object instance segmentation built upon FasterRCNN, which uses a branch of the network to predict a binary mask for each object instance. Top-down instance segmentation has also been accomplished using recurrent networks with attention mechanisms, either by extracting visual characteristics and producing instance labels one item at a time or by guiding the formation of bounding boxes followed by a segmentation network \citep{ren2017end, ghosh2019understanding}. The Flood Filling Network (FFN) uses this concept to obtain individual object masks directly from raw image pixels \citep{januszewski2018high} and has also been used for EM segmentation as reviewed below. 


The other approach is known as proposal-free, which aims to combine semantic and instance segmentation in a bottom-up approach. This was the strategy taken by \citet{chen2017dcan}, where the prediction of contours/edges of objects along with semantic masks were incorporated into FCNs in a multi-task learning approach. Both contour/edge maps and semantic masks were then fused to obtain the instance segmentation maps. Other approaches use boundary-aware instance information (e.g. the distance between object boundaries or the amount of overlap between objects) to fuse edge features with intermediate layers of the network \citep{bai2017deep,oda2018besnet}. 

Semantic instance segmentation is another family of techniques that addresses instance segmentation with semantic-based approaches. Instead of inferring for each pixel the probability of belonging to a certain class, they infer the probability of belonging to a certain instance of a class. In fact, \citet{de2017semantic} proposed a discriminative loss function in this regard and demonstrated that it is superior than the cross-entropy and Dice loss function for instance segmentation. The discriminative loss function consists of three terms: a segmentation term, which penalizes incorrect class predictions; a boundary term, which penalizes incorrect boundary predictions; and a regularization term, which encourages smoothness in the predicted masks.}

\section{Fully supervised methods}
\label{sec:supervised}
Fully supervised methods use annotated images (training data) to learn computational models that can segment structures in unseen images from similar distributions (test data). The training set is used by the algorithm to determine the model's parameters in such a way as to maximize the model's generalization ability. Table~\ref{tab:supervised} summarizes the 33 papers (of the 38) that have used supervised learning for the semantic and instance segmentation of (sub) cellular structures.

\subsection{End-to-end learning - semantic segmentation}

End-to-end learning is a machine learning approach where a single model learns to perform a task without relying on pre-defined intermediate steps or features. Instead, the model is trained to map the input data directly to the desired output, in a single end-to-end process. End-to-end learning has become increasingly popular in recent years due to advances in deep learning, which allow the creation of models with large numbers of layers that can learn complex representations of data. These models are trained using backpropagation, a method for updating the weights of the model based on the error generated by a given loss function between the predicted output and the true output, which allows the model to improve its performance over time.

The 16 papers that fall within this category are focused on the semantic segmentation of two main cellular structures, namely NM - neuronal membranes (8 papers) and M - mitochondria (5 papers). Other structures include N - nuclei, NE - nuclear envelopes, and L - lysosome. 

Neuronal membrane segmentation refers to the process of identifying and separating the neuronal membrane from other structures in an EM image. Segmenting neuronal membranes in EM volumes helps partition an image into distinct regions that represent different neuronal cells and processes. It is essential for studying the function of neurons along with their synaptic connections for understanding the different signaling pathways in the brain. Digital reconstruction or tracing of 3D neurons depends on the accuracy of neuronal membrane segmentation as discontinuities could lead to merge and split errors which in turn affect the reconstruction.  

Similarly, mitochondria segmentation is the process of identifying and separating mitochondria, a type of organelle found in eukaryotic cells, from other structures in an EM image. Mitochondria segmentation is a challenging task due to the variability in their size, shape, and distribution within cells. Accurately segmenting mitochondria in 2D and 3D is important for studying the structure and function of these organelles, as well as investigating their role in various diseases. 

Below we categorize the proposed approaches based on their underlying 2D or 3D CNN architectures.
\subsubsection{Approaches based on 2D CNNs}
\begin{table*}[h!]
\centering
\scriptsize
\captionsetup{width=1.3\textwidth}
  \caption{The list of 33 (out of 38) papers reviewed in this work that are based on fully supervised learning frameworks with 2D and 3D CNN architectures applied to both semantic and instance segmentation. The abbreviation Org. stands for the studied organelle/s. \RA{The Type (2D and/or 3D) column indicates the type of methods used and problems addressed. The studies that are marked as both 2D and 3D use a 2D backbone method coupled with some post-processing operations for 3D reconstruction. The other studies that are flagged as 2D or 3D only, use 2D or 3D only backbones to address 2D or 3D problems, respectively.} The numbers in the Datasets column serve as correspondences to the identifiers in Table~\ref{tab:datasets}, and the definitions of the performance metrics are presented in Section~\ref{sec:metrics}.}
    \label{tab:supervised}
    \hspace*{-1in}\begin{tabular}    {p{4cm}@{\hspace{2mm}}p{2cm}c@{\hspace{1mm}}c@{\hspace{2mm}}p{1.2cm}@{\hspace{5mm}}p{2.5cm}@{\hspace{2mm}}p{1.8cm}@{\hspace{2mm}}p{5.6cm}}
    \toprule
    \textbf{Citation} & 
    \textbf{Org.} & 
    \multicolumn{2}{l}{~\textbf{Type}} &    
    \textbf{Datasets} & 
    \textbf{Performance} & 
    \textbf{Backbone} & 
    \textbf{Main methodological component\/s} \\    
    && \textbf{2D} & \textbf{3D}& &\textbf{metrics}&&\\    
    \midrule \\
    \multicolumn{6}{l}{\textbf{End-to-end learning - semantic segmentation}} \\
    \midrule
\cite{fakhry2017residual}& NM &\checkmark & \checkmark&1, 3 &RE, WE, PE& 2D U-Net&Residual blocks,  deconvolutions\\    
\cite{oztel2017mitochondria}&M &\checkmark&\checkmark&1&Acc, P, R, F1, JI &2D FCN& Block processing, Z-filtering\\
\cite{chen2017convolutional} &MT, M, PM, V& \checkmark & & 22& No evaluation &2D FCN & A CNN architecture with four layers\\
\cite{xiao2018deep}&NM&\checkmark &\checkmark &1& $V_{rand}$, $V_{info}$ & 2D FCN & Residual blocks, multi-level features\\
\cite{casser2018fast}&M&\checkmark&&4, 9& Acc, P, R, JI&2D U-Net&Few parameters, light-weight model\\ 
\cite{jiang2019effective}&N&\checkmark&&Private$^*$&JI, Acc &2D FCN&Residual, atrous, multi-level fusion \\  
\cite{cao2020denseunet}&NM&\checkmark & &1&$V_{rand}$&2D U-Net&Dense blocks, summation-skip \\
\cite{quan2021fusionnet}&NM &  \checkmark& &1&$V_{rand}$, $V_{info}$ &2D U-Net&Residual, summation-skip, multi-stage\\
\cite{spiers2021deep}&NE& \checkmark&\checkmark &19 &P, R, F1&2D U-Net& Tri-axis prediction\\
\cite{cheng2017volume}&M& &\checkmark&8& P, R, JI &3D U-Net&  Factorised convolutions\\ 
\cite{lee2017superhuman}&NM&&\checkmark&3&RE&3D U-Net&3D graph affinity, hybrid 2D-3D, residual  \\ 
 \cite{xiao2018automatic} & M&&\checkmark&5, 8& JI, DSC&3D U-Net&Hybrid 2D-3D, residual, auxiliary supervision \\
\cite{funke2018large}&NM& &\checkmark& 2, 10, 15& $V_{info}$, CREMI &3D U-Net& 3D graph affinity prediction\\
\cite{heinrich2018synaptic}&NM&&\checkmark&2&CREMI&3D U-Net &Signed distance regression map, hybrid 2D-3D \\  
\cite{mekuvc2020automatic}&M,~L&&\checkmark&13&TNR, R, DSC&3D FCN & HighRes3DZMNet, zero-mean, residual/atrous \\ 
 \cite{heinrich2021whole}&Many&&\checkmark&10&DSC&3D U-Net&Multi-class segmentation\\
\cite{bailoni2022gasp}&NM&&\checkmark& 2&ARAND&3D U-Net & Signed 3D graph affinity prediction\\

  \midrule \\
\multicolumn{6}{l}{\textbf{End-to-end learning - instance segmentation}} \\
\midrule
\cite{liu2020automatic}&M&\checkmark&&8&Acc, P, R, JI, DSC&Mask-RCNN &Recursive network, multiple bounding boxes  \\
  \cite{yuan2021hive}&M&\checkmark&\checkmark&  4, 8& JI, DSC, AJI, PQ&2D U-Net& Hierarchical view ensemble module, multi-task\\
\cite{luo2021hierarchical} &M&\checkmark&&4, 8&  JI, DSC, AJI, PQ&2D U-Net& Residual blocks, two-stage, shape soft-labels\\
\cite{wei2020mitoem}&M&&\checkmark& 6, 8&JI, AP-75  &3D U-Net& Mask, contour prediction, watershed\\
\cite{abdollahzadeh2021deepacson}&A,~N&&\checkmark&17&$V_{info}$, ARAND &3D U-Net& Shape-based postprocessing\\
\cite{lin2021nucmm}&N&&\checkmark&7& AP-50, AP-75, AP&3D U-Net&Hybrid 2D-3D module, residual blocks\\
\cite{li2022advanced}&M&&\checkmark &6& JI, DSC, AP&3D FCN &Hybrid 2D-3D module, multi-scale\\
  \cite{mekuvc2022automatic}&M& &\checkmark&13&TPR, TNR, JI, DSC&3D U-Net&HighRes3DzNet, geodesic active contours\\
 \midrule \\
\multicolumn{6}{l}{\textbf{Ensemble learning - semantic segmentation}} \\
\midrule
\cite{zeng2017deepem3d}&NM & &\checkmark& 3&RE&3D U-Net & Hybrid 3D-2D,  residual/inception/atrous\\
\cite{haberl2018cdeep3m} & NM, M, N, V & &\checkmark&18, 21 &A, P, R, F-1 &3D U-Net & Hybrid 3D-2D,  residual/inception/atrous\\
\cite{guay2021dense}&C,~M,~GN & &\checkmark& 16&Mean JI& 3D U-Net &  Hybrid 2D-3D, spatial pyramids  \\
 \cite{khadangi2021stellar}&M&\checkmark&&12, 18&Acc, TPR, TNR, F1, JI, $V_{rand}$, $V_{info}$& 2D U-Net &  Ensemble of different networks\\
\midrule\\
 \multicolumn{6}{l}{\textbf{Transfer learning - semantic segmentation}} \\
\midrule
\cite{dietlmeier2019few}&M&\checkmark&&1&Acc, P, F1&VGG& Few shot, hypercolumn features, boosting \\
\cite{bermudez2018domain}&M&\checkmark&&Private$^*$&JI&2D U-Net&Deep domain adaptation, two-stream U-Net \\
 \midrule\\
 \multicolumn{6}{l}{\textbf{Configurable networks - semantic segmentation}} \\
\midrule
\cite{isensee2019nnu}&NM&\checkmark&\checkmark&2&Acc, P, F1 &2D, 3D U-Net& nn U-Net, self-configuring method \\
\cite{franco2022stable}&M&\checkmark&\checkmark&4, 8&JI&2D, 3D U-Net& Stable networks, blended output, $z$-filtering \\
    \bottomrule
    \end{tabular}    
    \justifying
    $^*$Private indicates that the dataset used is not publicly available.\vspace*{0.5cm}
\end{table*}

Successes of DL networks for segmentation in EM were achieved using 2D architectures with deep contextual networks. Those networks typically had FCN or U-Net as backbones. Deeper contextual networks have generally produced better 2D segmentations that mostly allowed doing away with multi-step post-processing for obtaining 2D segmentations and 3D reconstructions.


Residual Deconvolutional Networks (RDN) by \citet{fakhry2017residual} are based on a combination of residual connections, which allow for the efficient training of deep networks, and deconvolutional layers in the decoder of 2D U-Net, which help to recover spatial information lost during downsampling. The proposed method was evaluated on the ISBI 2012 (Drosophila VNC) and 2013 (SNEMI3D) benchmark datasets and compared to several state-of-the-art segmentation methods. The results demonstrated that RDNs were superior in terms of segmentation accuracy and required a simple post-processing step such as watershed to segment/reconstruct neural circuits. 

\citet{oztel2017mitochondria} proposed using a median filtering approach to incorporate 3D context for the reconstruction of mitochondria from the output of 2D segmentations. An FCN was used for delineating mitochondria from the background followed by median filtering along the $z$ direction in the volume of images. This $z$-filtering allows the removal of spurious strokes and the recovery of regions of interest when sufficient adjacent slices contain the missed component.

The deep contextual residual network (DCR) by \citet{xiao2018deep} is an extension of FCN with residual blocks and multi-scale feature fusion. They
used the \RB{summation based skip connections} which fuse high-level details from output of deconvolutions in the decoder and low-level information from ResNet encoder. The proposed post-processing method with a multi-cut approach and 3D contextual features proved important to reduce discontinuities (boundary splits or merges), which in turn helped to reduce false positives and false negatives in various 2D sections. DCR outperformed several state-of-the-art segmentation methods on the ISBI 2012 dataset. 

Advanced networks for different tasks may be too computationally demanding to run on affordable hardware, leading users to modify macro-level design aspects. Examples of such modifications include downsampling input images and reducing network size or depth to ensure compatibility with computer hardware constraints. \citet{casser2018fast} introduced a fast mitochondria segmentation method using a reduced number of layers and lightweight bilinear upsampling instead of transposed convolutions in the decoder of U-Net. Moreover, they introduced a novel data augmentation method that generates training samples on the fly by randomly applying spatial transformations to the original images, which leads to increased training efficiency and robustness to variations in image quality. The authors also incorporate a post-processing step based on $z$-filtering to reconstruct 3D mitochondria. The proposed approach was evaluated on several EM datasets and achieved state-of-the-art performance in terms of segmentation accuracy and speed. 

Data augmentation is a technique that is mostly used in machine learning and computer vision to increase the size and diversity of a training set. This is the case with most of the papers that are reviewed here. The process involves applying various transformations or modifications to the existing data in order to create new, but similar, instances of the data. It is particularly useful in cases where the size of the available dataset is limited, as it allows the model to learn from a larger and more diverse set of data without requiring additional data collection efforts. It can also help prevent overfitting and improve the robustness of the model by exposing it to a wider range of data variations. Moreover, test-time augmentation has also been proven effective to average out noise in predictions but at the cost of time complexity \citep{lee2017superhuman, zeng2017deepem3d, xiao2018automatic, yuan2021hive}. 

A residual encoder module with ASPP for multi-scale contextual feature integration was investigated by \citet{jiang2019effective}. The decoder module included the fusion of previous
low-level features and high-level features from the output of ASPP, followed by bi-linear upsampling to obtain the segmentation map. They achieved better performance compared to the baseline, U-Net, and Deeplab v3+ for the segmentation of cell bodies and cell nuclei.

The Dense-UNet model was proposed by \citet{cao2020denseunet} as an extension of the popular U-Net architecture that incorporates densely connected blocks within the U-Net's skip connections. The densely connected blocks help to improve gradient flow and feature reuse, which leads to better feature representation and higher segmentation accuracy. Besides its outstanding results on the ISBI 2012 challenge, the model turned out to be highly robust to variations in noises and artifacts of neuronal membrane images, requiring no further post-processing.

FusionNet is a fully residual U-Net architecture that combines different levels of feature representations by fusing the output of multiple sub-networks with different receptive fields. It includes a residual learning framework along with deconvolutional layers to improve the training convergence and segmentation accuracy. The study by \citet{quan2021fusionnet} showed that an integrated multi-stage refinement process using four concatenated FusionNet units
can effectively eliminate the requirement for any proofreading\footnote{Proofreading refers to the manual validation of segmented (manual or automatic) image data.} 

A novel data augmentation strategy was also proposed by \citet{spiers2021deep}, which simulates realistic variations in the EM images to improve the robustness of their 2D CNN for the semantic segmentation of nuclear envelopes. The proposed approach based on 2D U-Net achieved high segmentation accuracy and can be used to extract meaningful biological information from the segmented nuclear envelope, such as the distribution of nuclear pores. Their model was run on each axis after transposing the stack, and the resulting three orthogonal predictions were merged to produce the ultimate segmentation.

\citet{chen2017convolutional} used a 2D CNN with only four layers for the segmentation of membranes, mitochondria, vesicles, and microtubules in cryo-ET. The architecture of the CNN layers was optimized to capture a large context by utilizing $15 \times 15$ pixel kernels in the first two layers. This design allowed for the use of a single max-pooling layer to downsample the output to half the input resolution, which aids in distinguishing intricate details of structures such as single (vesicle, microtubule) or double membrane (plasma membrane, mitochondria). A CNN for each of the four structures was trained with a few sections of the tomogram containing structures of interest. Automated segmentation was required for subsequent sub-tomogram classification and averaging for the determination of in-situ structures for the molecular components of interest.

\subsubsection{Approaches based on 3D CNNs}
Similar to 2D deep architectures, a 3D CNN consists of multiple layers of filters, including convolutional, pooling, and activation layers, to learn spatial features from the input data. The filters scan the input volume at different locations and orientations to identify features that are relevant for segmentation. The key difference between 2D and 3D CNNs is the inclusion of an additional depth dimension in the input data. This allows the network to capture the spatial and depth relationships between adjacent slices in the volume. Due to the large amount of data and computational resources required for training 3D CNNs, such methods are typically used in high-end computing environments, such as specialized workstations or cloud computing platforms. Hybrid 2D-3D architectures have also been investigated that try to find the right trade-off between high computational demand and effectiveness. 

In this review, there are three approaches that adopted complete 3D CNN architectures in a fully supervised way. 
The first is the work by \citet{cheng2017volume} who proposed a 3D CNN for the segmentation of mitochondria in volumetric data. The authors also propose a novel data augmentation technique that uses stochastic sampling in the pooling layers to generate realistic variations in the feature space. In their thorough investigation, they conclude that the 3D CNNs outperform their 2D counterparts with a high statistical significance. The improvement was mainly attributable to the introduced augmentations as well as to the factorized convolutions which also permitted high efficiency, which was also proven useful in FIB-SEM (isotropic) volumes.

\cite{mekuvc2020automatic} also presented a 3D CNN-based method for the segmentation of mitochondria and endolysosomes in volumetric EM. The proposed method is based on the HighRes3DNet architecture, but it has the filters in the first layer constrained to having zero mean, and called it HighRes3DZMNet. The zero mean layer made the neural network robust to changes in the brightness of the volume inputs. The network is trained using the UroCell dataset for jointly segmenting mitochondria and endolysosomes due to similar morphologies of these biological structures. The method was also applied to segment mitochondria in the Lucchi++ dataset to achieve state-of-the-art segmentation results for FIB-SEM volumes.

\citet{heinrich2021whole} also relied on a 3D CNN for the segmentation of 35 organelle classes in cells from FIB-SEM volumes. The multi-channel 3D U-Net was trained on 28 volumes from the open-source OpenOrganelle collection covering four different cell types. They investigated how one segmentation model that is trained with samples of all 35 organelles compares with more specific models that are trained with subsets of semantically-related organelle classes, such as the endoplasmic reticulum (ER) and its associated structures, namely ER exit sites, ER membrane, and ER lumen. It turned out, that the single model that is trained by all classes outperforms the more specific ones. This is attributable to the richer diversity in the training set which resulted in a model with better generalization abilities.


Hybrid 2D-3D approaches were adopted for the segmentation of volume datasets in order to reduce the computational cost of 3D convolutions in certain layers and achieve better convergence. Their main application lies in the ability to segment anisotropic volumes for efficiently processing their 3D context. For instance, both anisotropic and isotropic EM volumes could be processed using hybrid 2D-3D network architectures that include $3 \times 3 \times 1$ convolutions instead of $3 \times 3 \times 3$ to modify them to 2D ones. \citet{xiao2018automatic} was the first to introduce a fully residual hybrid 2D-3D network with deep supervision to improve mitochondria segmentation. For reducing the number of parameters, 3D convolutions were used only in the first and last layers of a 3D U-Net. A deeply supervised strategy was proposed by injecting auxiliary branches into the initial layers of the decoder for avoiding the vanishing gradients problem. The complexity of the network allowed it to use a simple connected component analysis method for 3D reconstruction across both anisotropic and isotropic volume EM datasets.

\citet{lee2017superhuman} adapted the hybrid 2D-3D model of \citet{turaga2010convolutional} to predict 3D affinity maps for the segmentation of neuronal membranes in 3D volumes. The proposed CNN model incorporated multi-slice inputs along with long-range affinity-based auxiliary supervision in both the $z$- and $x-y$ directions. They utilized a hybrid 2D-3D U-Net for segmenting anisotropic volumes and post-processing with a simple mean-affinity agglomeration strategy for segmenting neuronal regions. The proposed affinity supervision simulates the use of boundary maps with different thicknesses in the DeepEM3D (Section~ \ref{sec:ensemble}), outperforming it in the SNEMI3D competition. 

A structured loss that favors high affinities between 3D voxels was used to obtain topologically correct segmentations by \citet{funke2018large}. The affinity predictions were accurate enough to be used with a simple agglomeration to efficiently segment both isotropic and anisotropic (CREMI, FIB, and SegEM) data, outperforming methods with more elaborate post-processing pipelines. \citet{bailoni2022gasp} used signed graphs to anticipate both attractive and repulsive forces among 3D voxels, enabling graph prediction through a 3D U-Net, in a manner similar to the method proposed by \citet{funke2018large}.  

Building on the concept of long-range affinities for boundary detection, \citet{heinrich2018synaptic} used neighboring context to predict voxel-wise distance maps through regression loss instead of probabilities. Those distance predictions, when thresholded, generated precise binary segmentations for synapses. Such distance prediction maps with simple thresholding allowed scaling the prediction at high-throughput speeds (3 megavoxels per second) for a full adult fly brain volume of 50 teravoxels in size.

\subsection{End-to-end learning - instance segmentation}
End-to-end learning approaches are also the most popular ones for instance segmentation, which requires the delineation of each instance within the same class of structures. This is particularly important for classes of structures that tend to be apposed with each other, such as mitochondria.

CNN-based methods for instance segmentation were grouped into two categories by \citet{wei2020mitoem}: top-down and bottom-up. Top-down methods typically utilize region proposal networks followed by precise delineation in each region.  Conversely, bottom-up approaches aim to predict a binary segmentation mask, an affinity map, or a binary mask with  instance boundary followed by several post-processing steps to distinguish instances. Due to the undefined scale of bounding boxes in EM images, bottom-up approaches have been the preferred methodology for 2D and 3D instance segmentation. 

\begin{sidewaysfigure}[t]
\centering
\footnotesize
\begin{tabular}{@{}c@{\hspace{2mm}}c@{\hspace{2mm}}c@{}}
\includegraphics[width=5.75cm,height=3.275cm]{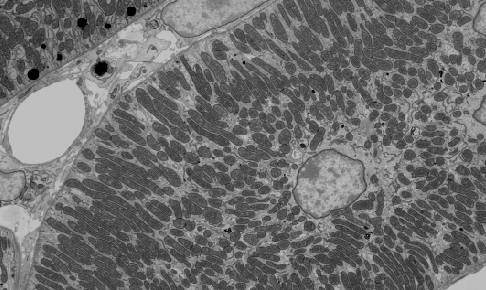} &
\includegraphics[width=5.75cm,height=3.275cm]{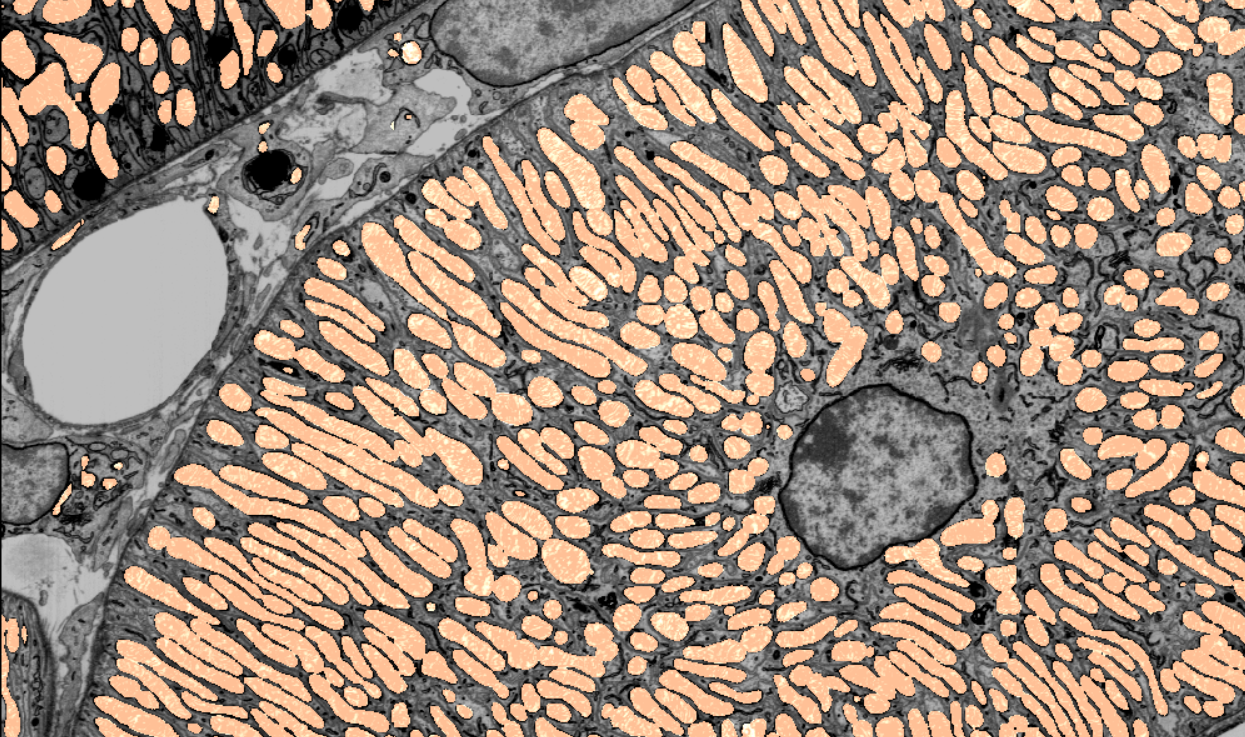} &
\includegraphics[width=5.75cm,height=3.275cm]{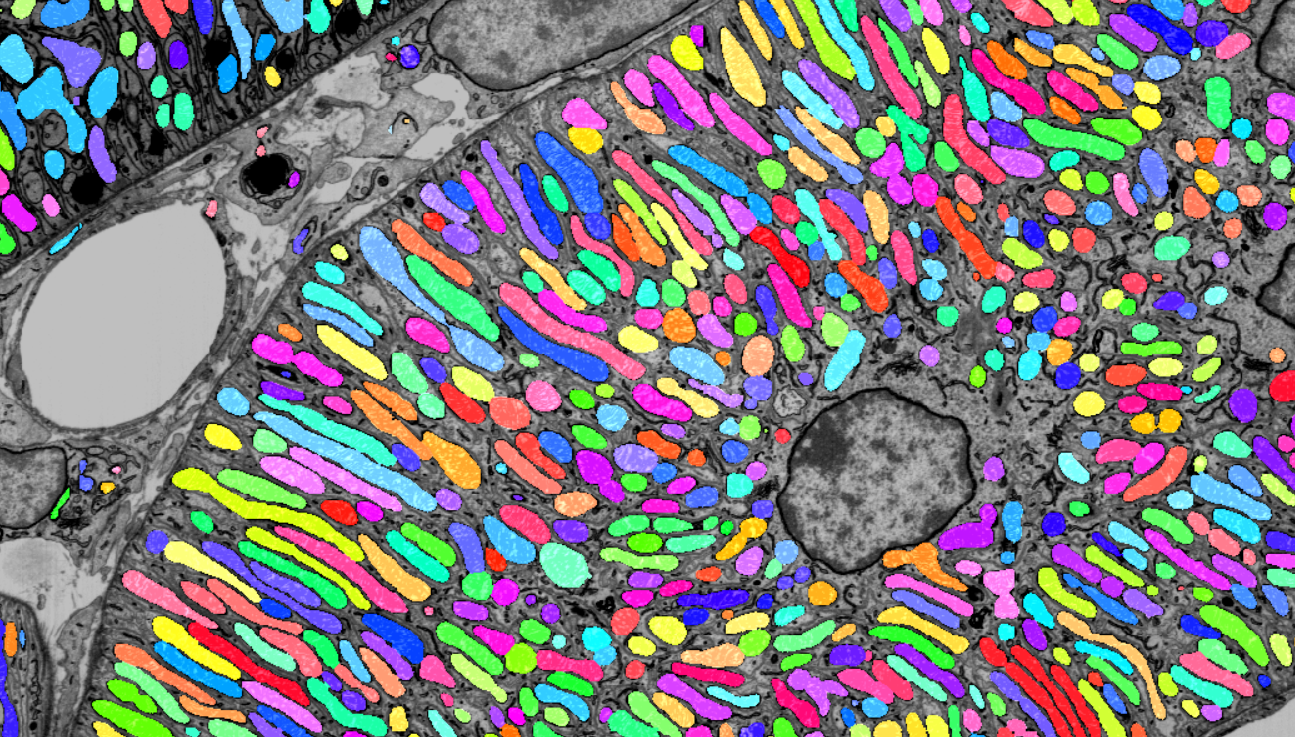}\\
(a) & (b) & (c) \vspace{2mm}\\
\includegraphics[width=5.75cm,height=3.275cm]{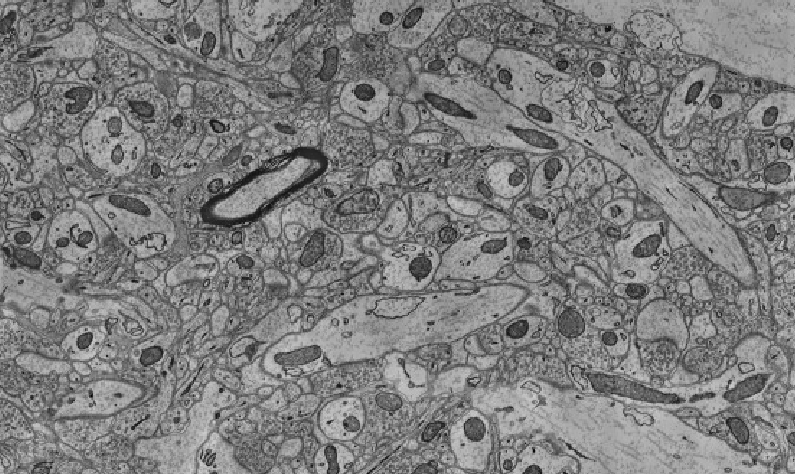} &
\includegraphics[width=5.75cm,height=3.275cm]{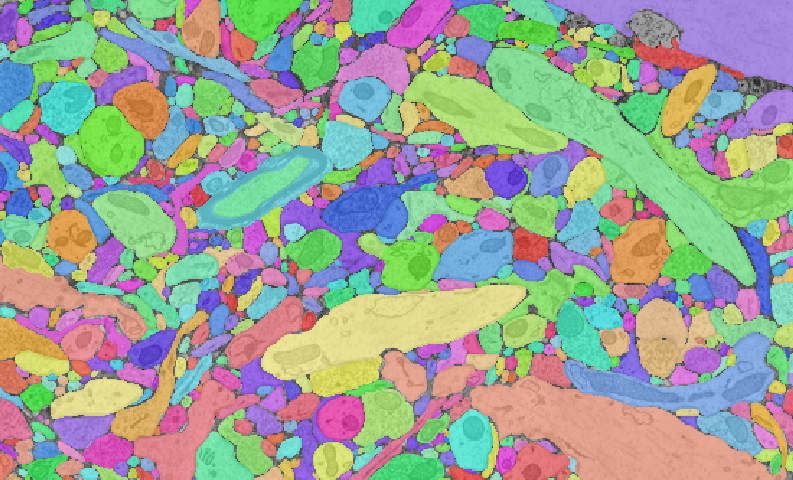} &
\includegraphics[width=5.75cm,height=3.275cm]{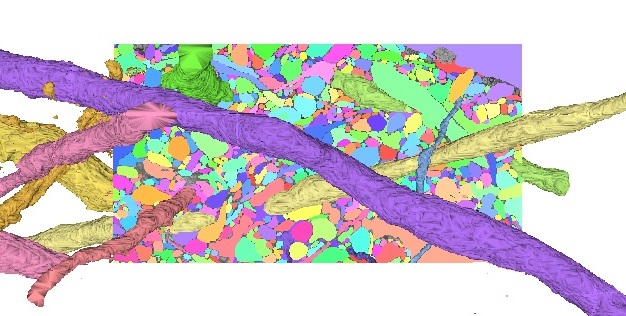} \\
(d) & (e) & (f) \\
\end{tabular}
\caption{Example of (top row) semantic and instance segmentation of mitochondria and (bottom row) neuronal membrane segmentation followed by 3D reconstruction of neuronal objects from a volumetric EM image. \textbf{(a)} Raw EM 2D section extracted from a FIB-SEM volume of a mouse kidney from the OpenOrganelle jrc\_mus-kidney dataset\protect\footnotemark. \textbf{(b, c)} Ground truth labels for semantic and instance segmentation. The instance segmentation map identifies each individual mitochondria with a unique color. \textbf{(d)} Raw EM 2D section extracted from the SNEMI3D (\#3) dataset for the task of neuronal membrane segmentation and reconstruction. \textbf{(e)} The ground truth map of the neuronal membrane segmentation, which is used to partition the image completely. \textbf{(f)} 3D reconstruction of selected neuronal structures that pass through the given 2D section from adjacent sections of the EM volume. The information from multiple images is used to create a 3D reconstruction through various post-processing methods, such as clustering, watershed, or graph-based methods.} 
\label{fig:semanticinstaceexample}
\end{sidewaysfigure}

The delineation of neuronal membrane does not require binary labels to distinguish one type of neuron from the other. There is no interesting semantics involved, unlike distinguishing a sub-cellular structure from other irrelevant structures or backgrounds followed by delineation to obtain individual instances. This type of segmentation is also referred to as image partitioning, as it divides the entire image into different neuronal parts based on its membranes. Such partitioning allows for the reconstruction of individual neuronal structures using post-processing. Figure~\ref{fig:semanticinstaceexample} shows examples of semantic and instance segmentation of mitochondria along with an illustration of neuronal 3D reconstruction after image partitioning.

\subsubsection{Approaches based on 2D CNNs}

The only top-down approach from the reviewed works in this paper is the one proposed by \citet{liu2020automatic}. They introduced a pipeline that complements Mask-RCNN. In particular, they proposed a  mechanism that refines undersegmented mitochondria in the output of Mask-RCNN, by iteratively enhancing the field of view that preserves the previous segmentation states. They systematically demonstrated that their approach outperformed competing methods that rely on U-Net, FFN, and Mask-RCNN in instance segmentation of mitochondria. 

\setcounter{footnote}{5}
\FloatBarrier
\FloatBarrier

\footnotetext{\url{https://open.quiltdata.com/b/janelia-cosem-datasets/tree/jrc_mus-kidney/}}




\RA{Shape prior turned out to be important for some techniques to improve the quality of instance segmentation. Shape prior refers to the incorporation of prior knowledge about the expected shape or structure of an object of interest into segmentation algorithms. 
For example, \citet{yuan2021hive} proposed the Hive-Net CNN, which was designed to overcome the challenges posed by the high variability in mitochondria shapes and sizes, as well as the presence of other cellular structures in the images. The network consists of multiple view-specific sub-networks that process different views of the image, and a centerline-aware hierarchical ensemble module that combines the outputs of the sub-networks to generate the final segmentation result. The centerline-aware module uses a new type of loss function that encourages the network to learn the morphology of mitochondria and to segment them along their centerlines. The proposed network was evaluated on two publicly available datasets, and an ablation study concluded that the centerline-aware module and the view-specific sub-networks were critical for achieving high segmentation accuracy. 

Shape information has also been exploited by the hierarchical encoder-decoder network (HED-Net) for the instance segmentation of mitochondria \citep{luo2021hierarchical}. That strategy leveraged the shape information available in the manual labels to train the model more effectively. Instead of relying solely on the ground truth label maps for model training, an additional subcategory-aware supervision was introduced. That was achieved by decomposing each manual label map into two complementary label maps based on the ovality of the mitochondria. The resulting three-label maps were used to supervise the training of the HED-Net. The original label map was used to guide the network to segment all mitochondria of varying shapes, while the auxiliary label maps guided the network to segment subcategories of mitochondria with circular and elliptic shapes, respectively. The experiments conducted on two publicly available benchmarks show that the proposed HED-Net outperforms state-of-the-art methods.

The inclusion of apriori knowledge about shape in segmentation algorithms contributes to increased specificity as they become more selective in delineating the structures of interest and keep false positives to a minimum. They can also improve generalization ability especially when the training data is limited. Methods that use shape priors, however, are more structure-specific, and therefore different methods may need to be designed for the segmentation of distinct organelles.}

\subsubsection{Approaches based on 3D CNNs}
The largest instance segmentation dataset for mitochondria (MitoEM) proposed by \citet{wei2020mitoem} benchmarks the dataset by proposing a 3D U-Net. It is trained with binary masks and contours using two separate decoders, followed by a marker-controlled watershed to obtain instance segmentations, and is called U3D-BC +MW for short. \citet{wei2020mitoem} introduced two networks, MitoEM-R and MitoEM-H, citing variations in sizes, shapes, and noise content for serial sections from rat and human samples. The MitoEM-R network can generalize on the human dataset as the rat samples have complex mitochondrial morphologies. The simpler U3D-BC +MW method was shown to be more effective than FFNs, as they were not able to capture the fine geometry of mitochondria with complex shapes or in close contact to each other. 

The DeepACSON approach by \citet{abdollahzadeh2021deepacson}, which was proposed for the instance segmentation of axons and nuclei in 3D volumes, is supported by a postprocessing method that relies on shape features. To correct for topological errors, a cylindrical shape decomposition algorithm is used as a postprocessing step to identify any erroneously detected axons and to correct under-segmented ones at their cross-overs. The circularity of the cell nucleus is corrected using the level-set based geometric deformable model, which approximates the initial shape of the object with a curve. This is then adjusted to minimize an energy function associated with the curve when it fits perfectly to the object’s boundaries. Energy functions enable the inclusion of shape information, whether it is a vague concept like smoothness constraints or a precise idea like shape constraints (strict adherence to a particular shape). 

Nuclei instance segmentation on a large-scale EM dataset was proposed by \citet{lin2021nucmm}. Their network, U3D-BCD, was inspired by the U3D-BC above but involved the additional learning of a signed Euclidean distance map along with foreground masks and instance contours to capture the structure of the background for segmentation. To locate the seeds for object centers, their methods starts by thresholding the predictions to identify markers with high foreground probability and distance value, but low contour probability. Next, the marker-controlled watershed transform algorithm is applied with the predicted distance map and seeds to generate masks. This approach has two advantages over the U3D-BC model \citep{wei2020mitoem}, which also utilizes marker-controlled watershed transform for decoding. Firstly, the consistency among the three representations is leveraged to locate the seeds, which is more robust than the U3D-BC method which relies on only two predictions. Secondly, it uses the smooth signed distance map in the watershed decoding process, which is more effective in capturing instance structure than the foreground probability map used in U3D-BC. 

\citet{li2022advanced} addressed 3D mitochondria instance segmentation with two supervised deep neural networks, namely ResUNet-H and ResU-Net-R, for the rat and human samples on the MitoEM dataset, respectively. Both networks produce outputs in the form of a semantic and instance boundary masks. Due to the increased difficulty of the human sample, Res-UNet-H has an additional decoder path to separately predict the semantic mask and instance boundary, while Res-UNet-R has only one path. Once the semantic mask and instance boundary are obtained, a seed map is synthesized, and the mitochondria instances are obtained using connected component labeling. To enhance the networks' segmentation performance, a simple but effective anisotropic convolution block is designed, and a multi-scale training strategy is deployed. The MitoEM dataset has sparsely distributed imaging noise, with the human sample having a stronger subjective noise level than the rat sample. To reduce the influence of noise on segmentation, an interpolation network was utilized to restore the regions with noise, which were coarsely marked by humans. Besides mitochondria instance segmentation, the proposed method was demonstrated to have superior performance for mitochondria semantic segmentation.

\citet{mekuvc2022automatic} extended their previous approach based on the HighRes3DZMNet by post-processing steps with active contours to separate apposing mitochondria and thus achieving instance segmentation. By means of experiments on the extended UroCell dataset they demonstrated that this new approach is more effective than the U3D-BC +MW method.

\subsection{Ensemble learning}
\label{sec:ensemble}
Ensemble learning methods combine outputs of multiple algorithms or models to obtain better predictive performance in terms of accuracy and generalization. Pixel- or voxel-wise averaging and the majority or median voting are among the main aggregation methods. 

An ensemble technique was in fact investigated by \citet{zeng2017deepem3d} for the segmentation of neuronal membranes in the brain volumes. They trained several variations of their DeepEM3D network, which could process different numbers of input slices and inputs with varying thicknesses of object boundaries. The DeepEM3D network extended the FCN architecture by introducing a hybrid network with 3D convolutions in the first two layers to enable integrating anisotropic information in the early stages, and 2D layers afterwards. DeepEM3D employed inception and residual modules, multiple dilated convolutions, and combined the result of three models that integrated one, three, and five consecutive serial sections. Employing an ensemble strategy for enhancing boundaries (by maximum superposition) within the probability maps generated by these models proved essential for performing with near-human accuracy in the SNEMI3D challenge.  

CDeep3M is a cloud implementation of DeepEM3D to segment various anisotropic SBF-SEM and cryo-ET datasets \citep{haberl2018cdeep3m}. Trained by a few sub-volumes of the cryo-ET tomogram, the resulting network was able to segment vesicles and membranes with high accuracy in other tomograms. The network implementation proved efficient for segmenting large-volume EM datasets such as SBF-SEM making it easier to analyze enormous amounts of imaging data.

The strengths of the ensemble paradigm was also confirmed by \citet{guay2021dense} for the segmentation of cytoplasm, mitochondria, and four types of granules in platelet cells. They demonstrated that the best segmentation performance (in terms of intersection over union) was achieved by combining the output of the top $k$ performing weak classifiers, with each such classifier learned by a small portion of the training data. Similar to above, each model was a hybrid 2D-3D network used to segment anisotropic SBF-SEM volumes. They also highlighted that besides its effectiveness, their ensemble paradigm ensured better reproducibility of the results in comparison to individual models that were sensitive to initialization.


Multiple network outputs were also combined with a workflow for binary EM segmentation provided by the EM-stellar platform \citep{khadangi2021stellar}. Unlike the above two approaches, \citet{khadangi2021stellar} used the ensemble paradigm to aggregate the output of different types of networks, namely CDeep3EM \citep{haberl18}, EM-Net \citep{khadangi2021net}, PReLU-Net \citep{he2015delving}, ResNet, SegNet, U-Net, and VGG-16. A cross-evaluation using a heatmap of different evaluation metrics revealed that no single deep architecture performs consistently well across all segmentation metrics. This is why ensemble approaches  have an edge over individual methods as they leverage the strengths of each underlying model as was demonstrated in the evaluation of two different datasets for mitochondria segmentation in cardiac and brain tissue.

\subsection{Transfer learning}
Transfer learning is a framework that adapts the knowledge acquired from one dataset to another, and is generally used when an application has an insufficient amount of training samples.  A pre-trained model is fine-tuned, usually in the final layers, with the training samples of a new dataset. This technique was used by \citet{mekuvc2020automatic} for the segmentation of mitochondria and endolysosomes from the background in EM images. Since mitochondria and endolysosomes share similar texture and mitochondria are more in abundance a binary segmentation model was first learned to segment mitochondria from the background. \RA{Subsequently, transfer learning was used to adapt the learned model for the segmentation of endolysosomes too. This was achieved by freezing all layers of the network except for the last one, which was fine-tuned by a smaller training set that included examples of endolysosomes. }This approach is a demonstration how transfer learning can be used when the availability of a certain structure is limited.


Fine-tuning a pre-trained network comes with the risk of overfitting to the few labeled training examples of the new dataset or application. This challenge has opened up new research avenues, namely few-shot learning and domain adaptation. The former can be a meta-learning approach that ``learns to learn" from a given pre-trained model when conditioned on a few training examples (referred to as the support set) to perform well on new queries passed through a fixed feature extractor \citep{shaban2017one}. 

Few-shot learning was the focus of the work by \citet{dietlmeier2019few}, who proposed a few-shot hypercolumn-based approach for mitochondria segmentation in cardiac and outer hair cells. The idea behind hypercolumn feature extraction was to extract features from different levels of a pre-trained CNN and combine them to form a single, high-dimensional feature representation for each pixel. The VGG-16 model pre-trained on the ImageNet dataset was used to extract hypercolumns, which were then passed through a linear regressor for actively selecting features. Only 20 labeled patches (2 $\%$- 98$\%$ train-test split) were used from a FIB-SEM stack for training a gradient-based boosting classifier (XGBoost). They showed how high segmentation accuracy on the Drosophila VNC dataset could be achieved by actively selecting features and learning using far less training data and even by using a single training sample (single-shot).


Domain adaptation is another form of transfer learning, where the source to target datasets share the same labels (classes) but have a different data distribution. Changes in data distribution can be due to slightly different experimental parameters during EM imaging or due to the imaging of different tissue types or body locations. 
\citet{bermudez2018domain} proposed the two-stream U-Net architecture, where the weights are related, yet different for each of the two domains, for supervised training on a few target labels. Only $10\%$ of labeled target data was required for domain adaptation to achieve state-of-the-art performance when compared to a U-Net trained on a fully annotated dataset. 

\subsection{Configurability and Reproducability}

\RA{A key challenge in designing CNNs is the determination of the right architecture for the problem at hand. This has motivated research effort in what are known self-configurable networks that can automatically determine certain design choices. A self-configurable network is thus a type of artificial neural network that is capable of dynamically adapting its structure and parameters based on the input data and task concerned. This concept was used by \citet{isensee2019nnu}, who proposed the no-new-Net (nnU-Net) framework that consists of a 2D U-Net, 3D U-Net and a cascade of two 3D U-Nets.~Self-configuration based on cross-validation was used to automatically determine some hyperparameters, such as the patch size, batch size and number of pooling operations. While it was shown to be very effective in various semantic segmentation problems in medical image benchmark datasets, its generalization ability in EM datasets has yet to be evaluated comprehensively. }

\begin{table*}[h!]
\centering
\footnotesize
\captionsetup{width=1.3\textwidth}
  \caption{The list of 5 (out of 38) papers reviewed in this work and that are based on semi-, un- and self-supervised learning frameworks along with co-relative light and electron microscopy (CLEM) as discussed in Section 4.2). The abbreviation Org. stands for the studied organelle/s. The Type (2D and/or 3D) column indicates the type of methods used and problems addressed. The studies that are marked as both 2D and 3D use a 2D backbone method coupled with some post-processing operations for 3D reconstruction. The other studies that are flagged as 2D or 3D only, use 2D or 3D only backbones to address 2D or 3D problems, respectively. The numbers in the Datasets column serve as correspondences to the identifiers in Table~\ref{tab:datasets}, and the definitions of the performance metrics are presented in Section~\ref{sec:metrics}.}
    \label{tab:summary_table}
    \hspace*{-1in}\begin{tabular}    {p{4cm}@{\hspace{2mm}}p{2cm}c@{\hspace{1mm}}c@{\hspace{2mm}}p{1.2cm}@{\hspace{5mm}}p{2.5cm}@{\hspace{2mm}}p{1.8cm}@{\hspace{2mm}}p{5.6cm}}    
    \toprule
    \textbf{Citation} & 
    \textbf{Org.} & 
    \multicolumn{2}{l}{~\textbf{Type}} &    
    \textbf{Datasets} & 
    \textbf{Performance} & 
    \textbf{Backbone} & 
    \textbf{Main methodological component\/s} \\    
    && \textbf{2D} & \textbf{3D}& &\textbf{metrics}&&\\    
    \midrule \\
    \multicolumn{8}{l}{\textbf{Semi-supervised learning - The superscripts $S$ and $I$ indicate semantic and instance segmentation}} \\
    \midrule
\cite{takaya2021sequential}$^S$&NM & $\checkmark$ & &1&$V_{rand}$, $V_{info}$ &FCN& Sequential semi-supervised learning\\
  \cite{wolny2022sparse}$^I$&M & \checkmark&&1,6&AP-50, AP&2D U-Net  & Positive unlabeled, momentum encoder \\
   \midrule \\
    \multicolumn{6}{l}{\textbf{Unsupervised learning - Semantic segmentation}} \\
    \midrule
  \cite{bermudez2019visual}&M, S&\checkmark&&1, 3, 8& JI &2D U-Net&Two stream U-Net, domain adaptation\\
  \cite{peng2020unsupervised}&M&\checkmark&&3, 8 
 & JI, DSC&2D U-Net& Domain discriminators for adversarial loss\\
    \midrule \\
    \multicolumn{6}{l}{\textbf{Self-supervised learning - Semantic segmentation}} \\
    \midrule
 \cite{conrad2021cem500k}&M&\checkmark&\checkmark&2, 4, 8, 10,13, 18&JI&3D U-Net&Self-supervised learning, fine-tuning\\
        \bottomrule
    \end{tabular}    
\end{table*}

An experimental study by \citet{franco2022stable} uncovered substantial reproducibility issues of different networks proposed for mitochondria segmentation in EM data. Additionally, it distinguished the impact of innovative architectures from that of training choices (such as pre-processing, data augmentation, output reconstruction, and post-processing strategies) by conducting multiple executions of the same configurations. Their systematic analysis enabled the identification of stable and lightweight models that consistently deliver state-of-the-art performance on publicly available datasets.

\section{Semi-, un- and self-supervised methods}
\label{sec:self_unsupervised}
Semi-supervised and unsupervised learning are two types of machine learning methods, whose main difference between them is the amount of labeled data they use to train the model.

Unsupervised learning is a type of machine learning that deals with finding patterns and relationships in unlabeled training data. In this case, the algorithm learns to identify patterns and relationships in the data by clustering or grouping similar data points together. Semi-supervised learning, on the other hand, is a combination of supervised and unsupervised learning. It uses both labeled and unlabeled data to train the model. The labeled data is used to train the model on specific tasks, while the unlabeled data is used to help the algorithm learn patterns and relationships in the data \citep{zhu2009introduction}. In self-supervised learning, a model is trained on a dataset with labels that are automatically generated from the data itself. The goal is to learn useful representations of the data that can be used for downstream tasks, such as segmentation. 

A common strategy for semi-supervised learning is to use label propagation through self-training. The process begins by training a classifier on labeled samples and then classifying the unlabeled samples. A selection of these samples based on an active selection strategy or learned classifier is then added to the training set and the process is repeated multiple times \citep{cheplygina18}. This can be performed either inductively or transductively. The former refers to training a model on unseen targets to add new information to the previously trained model so that it can generalize on new unseen data, and the latter to training a model based on a select subset of labeled and unlabeled data to be able to predict correctly on a limited set of seen targets. 

A semi-supervised approach was proposed by \citet{takaya2021sequential} for the segmentation of neuronal membranes. They called their approach 4S that stands for sequential semi-supervised segmentation. It was based on the fact that adjacent images in a volume are strongly correlated. The goal of their method is to have a model that can only generalize to the next few slices instead of to the whole volume. This was achieved by starting with a few labeled slices that are used to train the first model. Then, in an iterative approach the model was used to infer the segmentation maps of a small set of subsequent images and the resulting segmentation maps were used as pseudolabels to retrain the model. Label propagation from labeled to the available unlabeled data was performed by predicting pseudo labels on the subsequent sections which represent the same targets and whose predictions could be included in the next round of model training as ground truth labels. It allowed the training to weigh the most recent inputs heavily unlike transfer learning where the goal is to generalize well on all use cases of the unlabeled dataset.


Another semi-supervised method was introduced by \citet{wolny2022sparse} for the segmentation of mitochondria and neuronal membranes. In contrast to the above, their goal was to train a model with a few manually annotated images, which can generalize for the whole dataset. In particular, they used a training set with a combination of positive labeled data and unlabeled data of positive and negative instances. As there is no direct supervision on the unlabeled part of the image, an embedding consistency term was introduced by training two networks on different data-augmented versions of each pixel. This was coupled with a push-pull loss function that they proposed to enforce constraints between different instances. It was realized by using anchor projections in the embedding space of a point in each instance to derive a soft label based on the set of surrounding pixels in the projected space. The instance segmentation was then achieved by grouping the pixel embeddings. This semi-supervised method is notable for a good tradeoff between segmentation performance and effort in manual annotation.

Unsupervised learning was explored by \citet{bermudez2019visual}, who investigated the unsupervised domain adaptation strategy for mitochondria segmentation to demonstrate how a model trained on one brain structure (source: mouse striatum) could be adapted to another brain structure (target: mouse hippocampus). Labeled data was only available to train the model on the source dataset (striatum). Visual correspondences were then used to determine pivot locations in the target dataset to characterize regions of mitochondria or synapses. These locations were then aggregated through a voting scheme to construct a consensus heatmap, which guided their model adaptation in two ways: a) optimizing model parameters to ensure agreement between predictions and their sets of correspondences, or b) incorporating high-scoring regions of the heatmap as soft labels in other domain adaptation pipelines. These unsupervised techniques yielded high-quality segmentations on unannotated volumes for mitochondria and synapses, consistent with results obtained under full supervision, without the need for new annotation effort.


In the case of severe domain shifts such as from a FIB-SEM to an ssSEM dataset as investigated by \citet{peng2020unsupervised}, adversarial learning may be used for domain adaptation in different tissues of various species. Adversarial learning is a machine learning paradigm that trains a model with an adversarial loss function that encourages the model to learn domain-invariant features. \citet{peng2020unsupervised} combined the geometrical cues from annotated labels with visual cues latent in images of both the source and target domains using adversarial domain adaptive multi-task learning. Instead of manually-defined shape priors, they learned geometrical cues from the source domain through adversarial learning, while jointly learning domain-invariant and discriminative features. \RA{By doing so, the model learned features that were useful for both source and target domains, and could perform well on the target domain despite having only labeled data in the source domain.} The method was evaluated extensively on three benchmarks under various settings through ablations, parameter analysis, and comparisons, demonstrating its superior performance in segmentation accuracy and visual quality compared to state-of-the-art methods.

Contrastive learning is a self-supervised paradigm where a model is trained to learn useful representations of input data by contrasting similar and dissimilar samples. The basic idea is to take a set of positive pairs (e.g., two different augmentations of the same image) and a set of negative pairs (e.g., two images containing different types of objects), and train the model to assign higher similarity scores to positive pairs and lower similarity scores to negative pairs. This results in a model that captures the underlying structure of the data and can be used for downstream tasks like classification, object detection, and semantic segmentation. \citet{conrad2021cem500k} used contrastive learning, specifically moment contrast, \citep{he2020momentum}, to learn useful feature representation from the unlabeled CEM500K dataset followed by transfer learning on given datasets. The heterogeneity of CEM500k coupled with the unsupervised initialization of a segmentation model contributed to achieving state-of-the-art results on six benchmark datasets that concern different types of organelles.

\section{Segmentation evaluation metrics}
\label{sec:metrics}
Segmentation methods are evaluated by measuring the extent of overlap between the ground truth (GT) and prediction (PR) segmentation maps.

\begin{figure*}[t]
\centering
\includegraphics[width=0.99\textwidth,keepaspectratio]{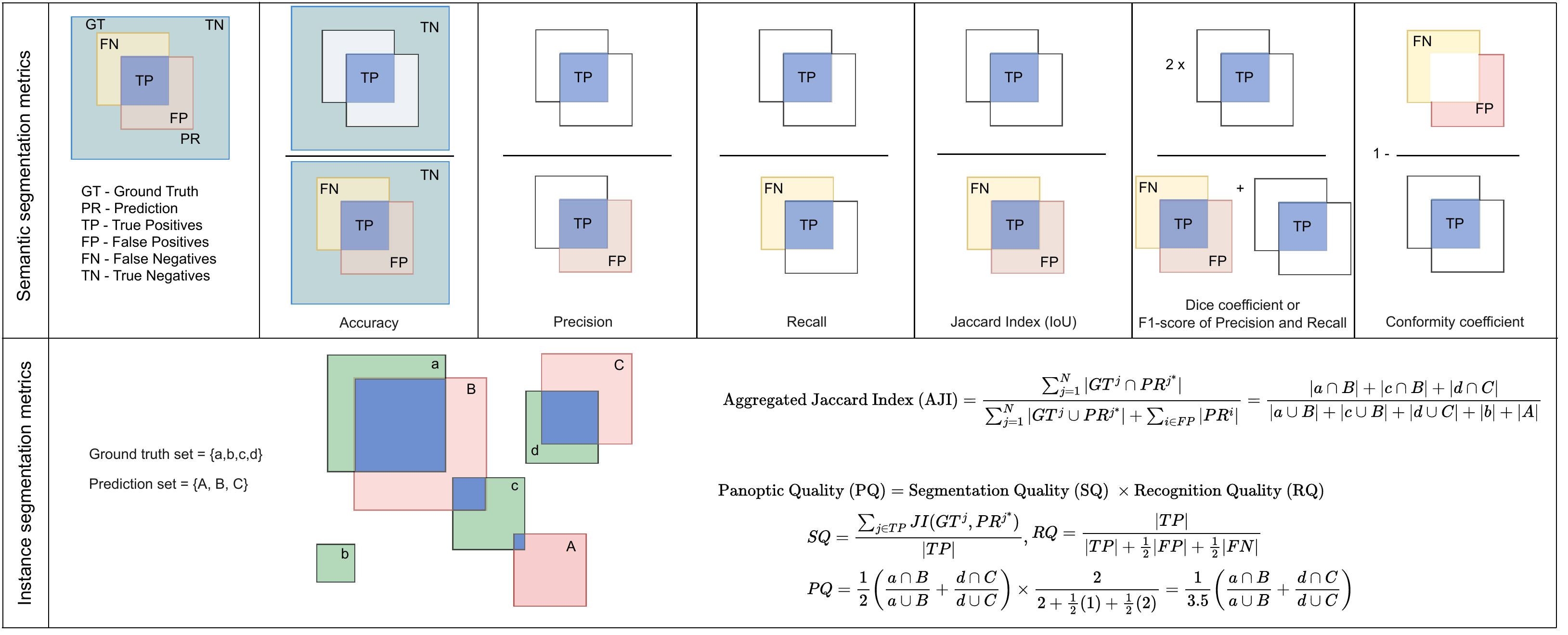}\vspace{-3mm}
\caption{Common performance metrics for segmentation methods. For semantic segmentation, the overall overlap of the ground truth (GT) mask with the prediction (PR) is compared without differentiating between objects of the foreground class. As to instance segmentation, each GT component is matched with only one PR component, the one with which it has the largest intersection. In the above example, the GT component `c' overlaps with two PR components, `A' and `B', but is matched only with `B' due to a larger overlap. The Aggregated Jaccard Index (AJI) is the ratio of the sum of all intersections of the matched pairs of GT and PR components to the sum of the unions of such pairs plus the sum of all pixels in the unmatched GT and PR components. The Panoptic Quality (PQ) captures both semantic and instance segmentation performance. The former is the sum of all IoUs between the matched GT and PR components divided by the number of matched components (TPs), and the latter is the number of TPs divided by the number of TPs plus half of the FPs and FNs together. The symbol $|.|$ indicates the cardinality of the set concerned.}
\label{fig:eval}
\end{figure*}

For semantic segmentation, all GT connected components are considered as one object, and similarly all PR connected components are treated as one object. This reduces the problem to binary classification. Typical performance measures include Accuracy, Precision and Recall and \RA{their harmonic mean (also called F1-score or Dice similarity coefficient (DSC)), the Jaccard Index (JI), also known as the Intersection over Union (IoU)}, and the Conformity coefficient \citet{chang2009performance}, Fig.~\ref{fig:eval}.

\RA{
\begin{equation}
    \begin{split}
    Accuracy &= (TP+TN)/(TP + FP +FN+TN)\\
    Precision~(P) &= TP/(TP+FP)\\
    Recall~(R) &= TP/(TP+FN)\\    
    F_1~(or~DSC) &= 2PR/(P+R) \\
    JI ~ (or ~ IoU) &= TP/ (TP + FP + FN)\\    
    Conformity &= 1-(FN+FP)/TP\\
    \end{split}
\end{equation}
}
\noindent where TP, FP, FN, and TN are the number of true positives, false positives, false negatives, and true negatives at pixel level. The Accuracy measure is a ratio of all correctly classified pixels to all pixels, Precision is the ratio of all true positive pixels to the number of positive predictions made by the algorithm, and Recall (Sensitivity or True Positive Rate) is the ratio of all true positive pixels to the number of all positive pixels in the ground truth. The JI (or IoU) and DSC measure the similarity between the predicted class labels and the true class labels, while the Conformity coefficient measures the ratio of the number of misclassified pixels to the number of true positive pixels subtracted from 1. A negative Conformity value indicates that the number of misclassified pixels is higher than the true positive ones, and vice-versa. Each of these metrics has its own strengths and weaknesses, and the choice of metric depends on the specific requirements of the classification task and the goals of the analysis. For example, accuracy is a simple and a good global measure but it is only suitable when the class distribution is balanced. 

For the segmentation of partitions, such as neuronal structures, the preservation of the topology is more important than the pixel-based accuracy. For instance, a prediction that oversegments (e.g. splitting the delineation of a neuron in two or more partitions) should be penalized more than a prediction of displaced, shrunk or expanded segments.~Metrics such as the Rand Index (RI), Warping error (WE), and variation of information (VOI) take into account the topological errors in neuronal membrane segmentation.~RI measures the similarity between the PR and GT segmentation maps, by calculating the sum of pairs of pixels that are both in the same object and both in different objects out of the total combination of pixel pairs in both GT and PR maps \citep{unnikrishnan2007toward, arbelaez2010contour}. The complement of the RI (i.e. 1 - RI) is known as the Rand Error (RE). The adapted rand error (ARAND) was the evaluation metric used in the SNEMI 3D challenge and is given as 1 - the maximal $F$-score of the RI. The maximal $F$-score is achieved when the precision and recall are at their optimal trade-off. The RI provides a score ranging from 0 to 1, with a value of 1 indicating perfect matching between two objects.

Other metrics that were part of the ISBI 2012 challenge are the WE and pixel error (PE). The WE is a segmentation metric that penalizes topological disagreements, i.e: the number of splits and mergers required to obtain the desired segmentation. On the other hand, the PE is defined as the ratio of the number of pixel locations at which the GT and PR labelings disagree. While expanding, shrinking, or translating a boundary between two neurons does not affect the WE, they incur a large PE. 

 The variation of information (VOI) quantifies the distance between PR and GT objects by measuring the amount of information that is lost or gained when one segmentation is transformed into the other \citep{arbelaez2010contour}. The VOI between the GT and PR components is the sum of two conditional entropies: the first one, $H(PR|GT)$, is a measure of over-segmentation, the second one, $H(GT|PR)$, a measure of under-segmentation. These measures are referred to as the VOI split or merge error, respectively. The VOI and ARAND were also combined to form the CREMI score by first taking the sum of the VOI split and VOI merge and then combining the result with ARAND using geometric mean. 

Evaluations of segmentation quality were most accurately reflected by the normalized versions of the RI and VOI, which are denoted by $V_{rand}$ and $V_{info}$, respectively \citep{arganda2015crowdsourcing}. Given a pair of GT and PR segmentation maps, $V_{rand}$ provides a weighted harmonic mean of the split and merge errors, where the split error is the probability of two selected pixels belonging to the same segment in PR given that they belong to the same segment in GT, and vice versa for the merge error. In pixel pair classification, the split and merge scores can be seen as representing precision and recall, respectively, for identifying whether the pixels belong to the same object (true positives) or different objects (true negatives). When the split and merge errors are weighed equally it is known as the Rand $F$-score. Similarly, $V_{info}$ is given by the mutual harmonic mean of the information-theoretic split and merge scores, which defines how much information in PR is provided by GT and vice versa. 


For the evaluation of object detection, where different connected components are treated as different objects, the above measures are also applicable. The main difference is the way a true positive is considered. In object detection a PR region is considered a TP if it overlaps with  more than a given threshold (e.g. 50\%) a GT component in terms of IoU, otherwise it is a FP. The unmatched GT components are then considered as FNs. A popular metric in object detection is average precision (AP), which is essentially the area under the precision-recall curve that is determined by systematically changing the detection threshold. The default AP measure uses a 50\% IoU overlap threshold, but other variations of the AP can be used depending on how strict the evaluation must be. The term AP-$\alpha$ denotes the average precision at a given IoU threshold $\alpha$. The higher the $\alpha$ the stricter the evaluation is. In problems with more than two classes, the mean AP (mAP) can be used to aggregate all the APs of all the classes involved by taking their average. 

Instance segmentation requires more detailed measures to quantify the segmentation mask accuracy along with the detection performance. Metrics such as the aggregated Jaccard index (AJI) and the Panoptic Quality (PQ), which were originally proposed by  \cite{kumar2017dataset} and \cite{ Kirillov_2019_CVPR}, respectively, have also been used in EM \citep{luo2021hierarchical, yuan2021hive} to evaluate instance segmentation algorithms more comprehensively. See Fig.~\ref{fig:eval} for an example.


\begin{equation}
    \begin{split}
    AJI &= \frac{\sum_{j=1}^N|GT^j\cap PR^{j^*}|}{\sum_{j=1}^N |GT^j \cup PR^{j^*}| + \sum_{i\in FP}|PR^i|}\\
    PQ &= \frac{\sum_{j \in TP}JI(GT^j,PR^{j^*})}{|TP|}\times \frac{|TP|}{|TP|+\frac{1}{2}|FP|+\frac{1}{2}|FN|}\\
    \end{split}
\end{equation}

\noindent where $N$ is the number of GT regions, and $j^{\ast}$ is the index of the connected region in $PR$ that is matched with the largest overlapping region (in terms of JI) with ground truth segment $GT^j$; FP is the set of false positive segments in $PR$ without the corresponding ground truth regions in $GT$, FN is the set of false negative segments in $GT$ that have been left unmatched with any regions in $PR$ and TP is the set of all matched regions in $GT$ and $PR$ with at least 50\% overlap in JI.
The symbol $|.|$ indicates the cardinality of a given set. A GT component can only be used once to match with a PR component. In case there are multiple PR components overlapping the same GT component, the GT component will only be matched with the PR component having the largest IoU. The AJI is an object-level performance metric which measures the ability of a segmentation algorithm to accurately identify and delineate individual objects within an image. It takes into account both the segmentation quality and the accuracy of object identification. For problems where many GT regions are apposing or in very close proximity with each other (e.g. mitochondria in 2D EM), there is a high risk that one PR region overlaps multiple GT regions. Such cases are overpenalised by the AJI measure. Overpenalization is prevented to happen with PQ because the matching of PR and GT regions are only valid when they overlap with more than 50\% in JI. 

\section{Discussion and open challenges}
\label{sec:disc}
Convolutional neural networks have become the standard choice for automatic feature extraction and segmentation in EM data. The most notable backbone networks are FCN and U-Net. Their use of deeper contextual network architectures is essential for accurate 2D prediction and by extension for 3D reconstruction. To produce dense predictions, early methods used a stack of successive convolutions followed by spatial pooling. Consecutive methods upsample high-level feature maps and combine them with low-level feature maps to restore crisp object boundaries and global information during decoding. To extend the receptive field of convolutions in the initial layers, numerous techniques have advocated the use of dilated or atrous convolutions. Recent works use spatial pyramid pooling to gather multi-scale contextual information in order to acquire global information in upper levels. More specialized architectures came into prominence to solve certain problems of anisotropy using hybrid 2D-3D networks and have now become the de facto for anisotropic EM datasets. The extension of 3D networks for graph-based affinity labeling proved useful as they can efficiently model structures across volume stacks to avoid several postprocessing steps for 3D reconstruction.
 
The advent of state-of-the-art deep neural networks led to a saturation of segmentation performance on small datasets, such as the ones from \citet{ciresan2012deep} and \citet{lucchi2011supervoxel}. Despite the focus in many studies on improving network architectures, a lot of the differences in performance can in fact be attributed to variations in preprocessing, data augmentation, and postprocessing \citep{isensee2019nnu,franco2022stable}. It is therefore likely that a focus on these areas in the near future will lead to new milestones in EM segmentation performance. Many of these developments, whether in network design or in postproceesing, have been in fully supervised segmentation, a technique that is greatly limited by the availability and quality of annotated data. While the number of large-scale annotated EM datasets has dramatically risen in recent years, the published datasets are not always precisely annotated, and are often composed of crude masks built semi-automatically using pre-trained networks and proofreading.

 This limiting scarcity of annotation in EM is due to the complexity of the produced images, their large scale, and the ways in which the annotations are obtained. Manual annotations can either be performed by one or a few domain experts, or they could be performed collaboratively by large groups in a crowd-sourced manner. Expert annotations are more accurate and time consuming. For instance, the large-scale connectomics project required extensive labeling and proofreading \citep{plaza2018analyzing}. Crowd-sourced approaches on the other hand require additional organizational efforts, specialized software, and instruction of the participants \citep{spiers2021deep}.

In addition to manual annotation, EM images can be labeled using specialized imaging modalities that target specific structures in the sample. For instance, CLEM (Correlative light electron microscopy) is used to label structures targeted with fluorescent probes at (sub)cellular scales \citep{de2015correlated,drawitsch2018fluoem,heinrich2021whole}. Other EM modalities that could assist in annotation include energy dispersive X-ray spectroscopy (EDX), electron energy loss spectroscopy (EELS), cathodoluminescence (CL), and secondary ion mass spectroscopy at the nanoscale (NanoSIMS) \citep{pirozzi2018colorem}. These methods reduce the bias in human annotation, but may require longer sample preparation, specialized equipment, or additional image processing to produce segmentations.

The scarcity of annotated EM data could also be addressed algorithmically by relying on semi-supervised and unsupervised learning techniques. These techniques are able to segment EM images with minimal or no annotations, that can scale to larger datasets with varied structures. Few-shot learning for segmentation has shown promising results in natural images \citep{dong2018few, tao2020few}, and the use of transformers in segmentation could prove useful for large-scale EM data in the future \citep{dosovitskiy2020image, zheng2021rethinking}.
 
These prospects of label-free segmentation highlight the importance of collecting unlabeled yet relevant segmentation datasets, like the curation of unlabeled heterogeneous mitochondria images in CEM500K. Such datasets have shown how unlabeled pre-training using self-supervision could pave the way for breakthroughs in EM segmentation.

Complex datasets with challenges such as MitoEM, NucMM, or even unlabeled datasets such as CEM500K led to deep generalist models rather than specialized networks but still have a long way to go as sub-cellular image segmentation using large-scale EM is yet to explore both challenges in biological research and computer vision.

\section{Acknowledgement}
This project has received funding from the Centre for Data Science and Systems Complexity at the University of Groningen\footnote{\url{www.rug.nl/research/fse/themes/dssc/}}. Part of the work has been sponsored by ZonMW grant 91111.006; the Netherlands Electron Microscopy Infrastructure (NEMI), NWO National Roadmap for Large-Scale Research Infrastructure of the Dutch Research Council (NWO 184.034.014); the Network for Pancreatic Organ donors with Diabetes (nPOD; RRID:SCR$_014641$), a collaborative T1D research project sponsored by JDRF (nPOD: $5-SRA-2018-557-Q-R$) and The Leona M. \& Harry B. Helmsley Charitable Trust (Grant $2018PG-T1D053$). The content and views expressed are the responsibility of the authors and do not necessarily reflect the official view of nPOD. Organ Procurement Organizations (OPO) partnering with nPOD to provide research resources are listed in \url{http://www.jdrfnpod.org/for-partners/npod-partners/}. Thanks are also due to Kim Kats for her assistance in preparing Fig. 1.

\bibliographystyle{model2-names.bst}\biboptions{authoryear}
\bibliography{sample.bib}
\end{document}